%% file: main.tex
\documentclass[10pt]{article}
\usepackage[letterpaper,margin=1in]{geometry}

\usepackage{lineno}

\usepackage{times}
\usepackage{latexsym}

\usepackage{booktabs} 
\usepackage{changepage,threeparttable}

\usepackage[utf8]{inputenc} 

\usepackage{microtype} 
\usepackage{inconsolata}

\usepackage[most]{tcolorbox}
\usepackage{xcolor, soul}  
\usepackage{color}
\usepackage[dvipsnames]{xcolor}
  
\usepackage[utf8]{inputenc} 
\usepackage[T1]{fontenc}    
\usepackage{hyperref}       
\usepackage{amsfonts,amsmath,amssymb,amsthm}       
\usepackage{array}
\usepackage{url}            
\usepackage{booktabs}       
\usepackage{nicefrac}       
\usepackage{microtype}      
\usepackage{bm}

\usepackage{booktabs,multirow,array}
\renewcommand{\arraystretch}{1.5}
\newcolumntype{C}[1]{>{\centering\let\newline\\\arraybackslash\hspace{0pt}}m{#1}}

\newcolumntype{N}{@{}m{0pt}@{}}
\usepackage{hhline} 

\usepackage{natbib} 
\usepackage{graphicx}
\usepackage{caption}
\usepackage{subcaption}
\usepackage{hyperref}

\usepackage{bigints}
\usepackage{amssymb,amsopn,algorithm,algorithmic,float,bbm,bm,enumerate,color,multirow,gensymb}

\usepackage{epsfig,graphicx}
\usepackage{comment}
\usepackage{afterpage}
\usepackage{thmtools,thm-restate}

\input{notation}

\usepackage[suppress]{color-edits}
\addauthor{ab}{blue}
\addauthor{abb}{magenta}
\addauthor{pc}{red}
\addauthor{sk}{green}

\usepackage{microtype}

\usepackage{times}

\usepackage{booktabs} 
\usepackage{changepage,threeparttable} 

\usepackage{arydshln}

\usepackage{nicematrix}
\usepackage{tabularx}

\title{Large Language Models can Achieve Social Balance}

\author{Pedro Cisneros-Velarde \\
  VMware Research \\
  \texttt{pacisne@gmail.com} \\}

\definecolor{lightblue}{RGB}{212, 235, 255}
\definecolor{grey1}{RGB}{96, 101, 102}
\definecolor{lightorange}{RGB}{255, 204, 168}
\definecolor{lightyellow}{RGB}{255, 255, 168}
\definecolor{lightgreen}{RGB}{224, 242, 213}
\definecolor{lightred}{RGB}{249,202,202}
\definecolor{lightgray}{RGB}{230,230,230}
\definecolor{deepred}{RGB}{152, 1, 0}
\definecolor{deepblue}{RGB}{41, 90, 168}
\definecolor{deeppink}{RGB}{238, 123, 145}

\definecolor{lightpink}{RGB}{255, 218, 218}

\newcommand{\lightp}[1]{\sethlcolor{lightpink}\hl{#1}}

\renewcommand{\arraystretch}{0.9} 

\usepackage{anyfontsize}

\date{}

\begin{document}
%
\maketitle

\begin{abstract}
Large Language Models (LLMs) can be deployed in situations where they process positive/negative interactions with other agents. We study how this is done under the sociological framework of social balance, which explains the emergence of one faction or multiple antagonistic ones among agents. Across different LLM models, we find that balance depends on the (i) type of interaction, (ii) update mechanism, and (iii) population size. Across (i)-(iii), we characterize the frequency at which social balance is achieved, the justifications for the social dynamics, and the diversity and stability of interactions. Finally, we explain how our findings inform the deployment of agentic systems.
\end{abstract}

\section{Introduction}
\label{sec:arXiv_intro}
\input{sec/intro}


\section{Problem Setting}
\label{sec:prob-sett}
\input{sec/dlopt_resumido}

\section{Analysis: Single Triad Population}
\label{sec:triad}
\input{sec/triad}

\section{Analysis: Multiple Triads Population}
\label{sec:mult}
\input{sec/multiple}

%

\bibliography{biblio}

\appendix

\section{Further Motivation for Our Work}
\label{app:furthmotiv-0}
\input{app/further-motiv_0}

\section{Related Literature}
\label{sec:arXiv_related}
\input{app/rel_works}

\section{Additional Information for Section~\ref{sec:triad}}
\label{app:more-sec3}

\input{app/more-info-sec3}

\section{Informing an Application on Social Media}
\label{sec:disc}
\input{sec/disc}

\section{Further Discussion on Our Work}
\label{app:further-motiv}
\input{app/further-motiv}
\subsection{Motivating our choice of how social information is shown to LLMs}
\label{app:motivchoiceinfo}
\input{app/choiceinfo}

\section{Implementation Details}
\label{app:exp-details}

\subsection{Problem setting}
\label{sec:prob-sett-exp}
\input{sec/dlopt}

\input{app/experimental_details}


\section{Keyword Analysis for Single Triad Population}
\label{sub:ka}
\input{app/qr-st}

\end{document}

%% file: notation.tex
\newcommand{\beq}{\begin{equation}}
\newcommand{\eeq}{\end{equation}}

\theoremstyle{definition}

\newcommand {\commentout}[1] {}

\def\ints{{{\rm Z} \kern -.35em {\rm Z} }}  
\def\smallints{{{\rm Z} \kern -.3em {\rm Z} }}  
\def\pints{{{\rm I} \kern -.15em {\rm N} }}      
\newcommand{\reals}{\mathbb R}

\def\cplx{{{\rm I} \kern -.45em {\rm C} }}       
\def\l2{\rm {\mathcal L}^{2}(\reals)}            

\newtheorem{nad}{Notation and Definitions}[section]

\newcommand{\be}{\begin{eqnarray}}
\newcommand{\ee}{\end{eqnarray}}
\newcommand{\bea}{\begin{eqnarray}}
\newcommand{\eea}{\end{eqnarray}}
\newcommand{\beaa}{\begin{eqnarray*}}
\newcommand{\eeaa}{\end{eqnarray*}}
\newcommand{\bnad}{\begin{nad}}
\newcommand{\enad}{\end{nad}}



%% file: sec/intro.tex
Large Language Models (LLMs) have shown language understanding~\citep{xi-2023-risellmagents,wei-2022-finetunedzeroshot,wei-2022-emergent} and reasoning~\citep{yao-2023-treeofthoughts,bang-etal-2023-multitask} abilities. 
In multi-agent settings, they can express complex opinion dynamics~\citep{chuang-2024-simulating,chuang-2024-wisdom,cisnerosvelarde2024princopindynLLM}. 
These abilities are influenced by biases~\citep{liang-2023-holistic}
from pre-training 
and/or alignment. 
%
%
These biases also influence how LLMs 
display sociopsychological qualities~\citep{zhou2024sotopia}, 
express~\citep{serapiogarcia2023personalitytraitslargelanguage} or assess~\citep{rao2023can} personality traits 
and
values or judgment~\citep{wang-etal-2024-incharacter,sorensen2024valuekaleidoscope}.

%

%
%
%
%
%

Given the abilities and biased behavior of LLMs, it is relevant to understand how they assess positive and negative interactions 
for applications 
where their responses or actions 
depend on such assessments
~\citep{guo-2024-llmmultiag}. For example, LLMs can participate in 
social media~\citep{leskovec-2010-SignedNetSocMedia}, where establishing negative ties have great influence on opinion mobilization~\citep{SrijanHamilton-2018-ComInterConflWeb}; 
negative ties are also 
associated to information diffusion and centrality of individuals~\citep{ghasemianchristakis-2024-antagonisticties}. 
Section~\ref{sec:disc-main} further explains the application to social media. 
%
%
%
Thus, contributing to the 
literature on multi-agent LLMs in sociologically relevant scenarios~\citep{llmsum-2024-surv}, we study the 
question:
\emph{how do LLM agents understand and update positive and negative interactions among themselves?}
%
%
%
To the best of our knowledge, 
this question 
is 
underexplored.
%
%

\begin{figure}[t]
\centering
\includegraphics[width=0.45\columnwidth]{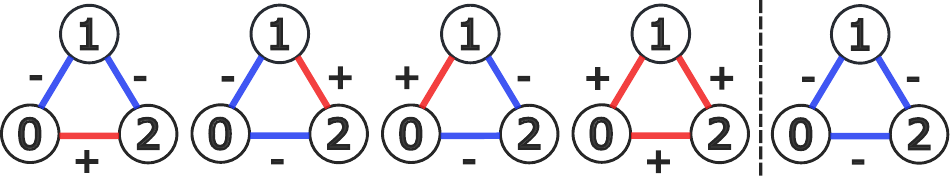}
\caption{\textbf{Sign configurations of a  balanced triad.}
Consider a triad as a graph with three nodes corresponding to the agents $0$, $1$, and $2$. 
The first four triads define structural balance (e.g., in the first triad: agent $2$ is an enemy of $1$, who in turn is an enemy of $0$; therefore, $2$ is a friend of $0$).
All five triads define 
clustering balance.
}
\label{fig:triads-signs}
\end{figure}

To study this problem concretely, 
we consider the 
well-established
sociological 
framework of \emph{social balance}. Social balance dictates the positive and negative interactions that must be held within 
\emph{every} 
three individuals or \emph{triad} in a population (e.g., a social network) 
for the emergence of \emph{factions}.
In particular, \emph{structural balance}
~\citep{heider-1946-AttCogOrg} 
%
characterizes the \emph{signed} interactions that lead to the emergence of one \emph{faction} where all individuals have positive interactions with each other, or two factions where 
interactions are only negative across factions.
%
Within a triad, structural balance is dictated by 
so-called 
``Heider's rules'':
(i) the enemy of an enemy is a friend; (ii) the friend of an enemy is an enemy; (iii) the enemy of a friend is an enemy; (iv) the friend of a friend is a friend.  
It is argued that when one of these rules is violated, it generates 
a \emph{cognitive dissonance} which the individual 
%
must resolve~\citep{festinger-1975-cogdis,FrancoisRic-2015-socialcognition}.
%
%
%
%
%
Another notion of social balance is \emph{clustering balance}~\citep{davis-1967-cluststructbalgraphs}, 
which happens when rule (i) is relaxed within a triad of agents, i.e., three agents can be enemies 
of 
each other. 
This allows 
the emergence of more than two antagonistic factions. 
Fig.~\ref{fig:triads-signs} 
illustrates 
all balanced triads.

Thus, in this paper, 
our main 
focus 
is to 
study the appearance of social balance in a group of \emph{three} LLM agents continuously updating their interactions with each other.
It is logical to first establish results for this fundamental case 
because a triad \emph{is} the basis of Heider's rules.
After this, we also study populations of multiple triads (more than three agents).
%
%
%
%

%
%

We characterize the interactions along two dimensions. 
The first 
is the \emph{type of interaction}: 
friendly and unfriendly ties among agents can be expressed as positive and negative \emph{relationships} with each other, or \emph{appraisals} or \emph{opinions} of each other. These
types 
are present in both theoretical and empirical sociological literature; e.g.,~\citep{SzellLambiotte-2010-MuiltOrgLScaleSocOn,leskovec-2010-SignedNetSocMedia,cisnerosvelarde-2020-SignFormClustering}. 
%
%
The second dimension is the \emph{update mechanism} for the agents' interactions. Motivated by mathematical sociology~\citep{mei-2019-DynSocBal}, we consider 
\emph{homophily} and \emph{influence}.
Assume Agent~$i$ wants to update its tie to Agent~$j$. 
Homophily is when Agent~$i$ does a comparison between (a) how it and Agent~$j$ relate to the other agents, and (b) how it directly relates to Agent~$j$. 
Influence is when Agent~$i$ considers, instead of (b), how Agent~$j$ directly relates to 
Agent~$i$.

Our \textbf{main contribution} is to show that diverse LLM models---of different sizes and family, having alignment or not, being a reasoning model or not---
are able to achieve social balance after repeated interactions: showing the emergence of one faction or multiple antagonistic ones in a social group.

\noindent\textbf{For a population of a single triad:} 
    
    The \emph{type} of social balance (structural or clustering), \emph{diversity} of \emph{sign configurations} of the achieved 
    balanced triads (Fig.~\ref{fig:triads-signs}), and
    \emph{frequency} of achieving social balance 
    differ across LLM models and types of interactions and update mechanisms.
    %
    %
    Generally, larger models achieve social balance more frequently and in its \emph{strongest} (structural) type. 
    %
    Heavily aligned models are biased toward either all-positive or all-negative interactions. 
    %
    %

    LLM models process update mechanisms differently: the \emph{same} update mechanism can have different and even opposite effects on the number of positive/negative interactions and on the diversity 
    of 
    balanced triads.
    %
    
    Remarkably, 
    %
    a model that achieves social balance more often does not necessarily have more  diversity of 
    socially 
    balanced triads or 
    more \emph{stable} (i.e., less changing) interactions. 
    %

    Finally, LLMs 
    show knowledge 
    of sociopsychological processes that lead to social balance: 
    %
    the 
    keyword ``dissonance'' 
    can be 
    used to justify 
    new 
    interactions that 
    \emph{do not contradict} Heider's rules---thus
    agreeing with 
    sociological theory. 
    A 
    reasoning model approaches the 
    update of its interactions 
    more mathematically.     
    \qed
    %
    %

\noindent\textbf{For a population of multiple triads:}

%
    Unlike the single triad case, 
    smaller models 
    achieve social balance more frequently 
    across most settings. 
    %
    Thus, achieving social balance is not a property that \emph{absolutely} improves with model size. 
    %
    Moreover,
    %
    more models achieve clustering balance.
    All of this is possible evidence that social interactions are updated under \emph{different} dynamics in larger populations. 
    Finally, the keyword ``dissonance'' is largely absent across LLM responses. \qed
    %
    %
    %
    %
%


We further discuss our work's motivation in 
App.~\ref{app:furthmotiv-0}. 
The related literature 
is in App.~\ref{sec:arXiv_related}.

%% file: sec/dlopt_resumido.tex
\begin{figure}[t!]
\centering
  \includegraphics[width=0.55\linewidth]{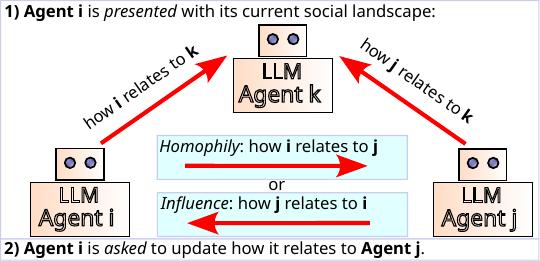}
  \caption{
  \textbf{Agent $i$ updates its interaction toward 
  Agent $j$.} 
  The update depends on how 
  both relate to every other Agent $k$ in the population.}
  %
  \label{fig:update}
\end{figure}

We consider a population of LLM agents of the same model, type of interaction 
(relationship, appraisal, or opinion), 
and update mechanism (homophily or influence). 
%
Initially at $t=0$, every agent has positive or negative interactions.
Interactions are directional and need not be 
symmetrical 
(e.g., Agent $i$ has a positive opinion of $j$, while $j$ has a negative one of $i$). 
For a single triad population, we initialize the interactions according to all $2^6$ possible sign distributions. 
We run ten simulations per sign distribution.
For a 
population size larger than three, 
we do ten random initialization for the interaction signs. 
%
At every iteration 
$t=1,\dots,10$, regardless of the population size, all agents simultaneously update their interactions toward every other agent; 
see 
Fig.~\ref{fig:update}. 
All agents are asked to justify the update of their interactions. 
A population achieves social balance when every triad in it satisfies Fig.~\ref{fig:triads-signs} 
(where all 
interactions are symmetrical) at iteration $t=10$. 
%
A more formal description 
and further experimental and prompt details are in App.~\ref{app:exp-details}.

%% file: sec/triad.tex
\begin{table}[t!]
    \centering
    \def\arraystretch{1}
    
     \small
    \resizebox{0.6\linewidth}{!}{
    \begin{tabularx}{0.6\linewidth}{c*{6}{>{\centering\arraybackslash}X}c}
    \toprule
    & \multicolumn{2}{c}{\textbf{Rel.}} & \multicolumn{2}{c}{\textbf{Apprais.}} & \multicolumn{2}{c}{\textbf{Opin.}} & \\
    \cmidrule{2-7}
    & \textbf{H} & \textbf{I} & \textbf{H} & \textbf{I} & \textbf{H} & \textbf{I} & 
    \textbf{type}
\\
    \midrule   
    \textbf{gpt-oss-120b} 
    & \underline{66.88} & \underline{36.72} & \underline{96.88} & 26.41 & \underline{95.31} & 23.75 & Both
    \\
    \textbf{Llama~3~70B} & \textbf{76.56} & \textbf{87.50} & \textbf{100.00} & \textbf{54.69} & 18.75 & \underline{26.88} & S\\
    \textbf{Llama~3~8B} & 5.63 & 25.62 & 44.22 & 0.00 & \textbf{98.44} & \textbf{56.25} & Both\\
    \textbf{Mistral~7B} & --- & 9.38 & 23.44 & 7.81 & 10.94 & 7.81 & C\\ 
    \textbf{Gemma~3~4B} 
    & 52.81 & 34.38 & 43.75 & \underline{27.50} & 27.66 & 15.63 & C
    \\
    \bottomrule
    \end{tabularx}
    }
    %
    %
     \caption{
    \textbf{Frequency (percentage \%) of simulations that result in social balance.}
    H: homophily, I: influence.
    We indicate whether each model has \emph{every} setting resulting \emph{exclusively} in structural (S)/clustering (C) balance, or if both types appear. Only Mistral~7B refuses to provide an answer under relationships with homophily.}
    \label{tab:single-triad-freq}
\end{table} 

\begin{figure}[t!]
\centering
  \includegraphics[width=0.60\linewidth]{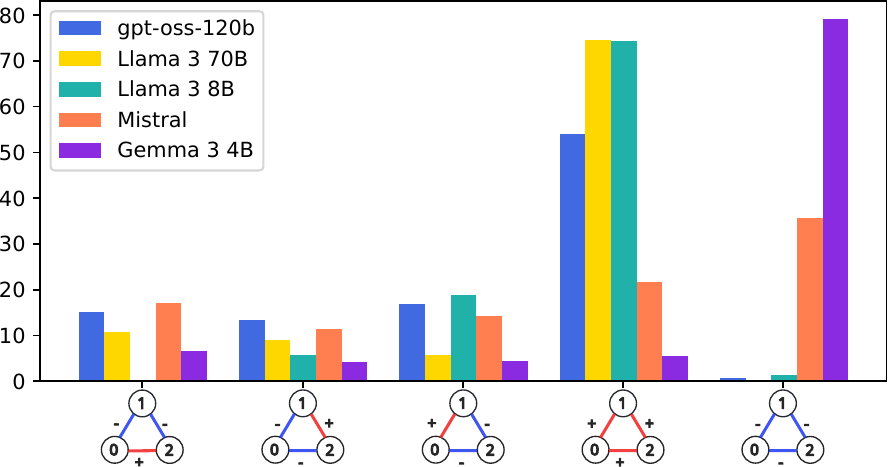}
  \caption{
  \textbf{Frequency (\%) 
  of balanced triads.} 
  For each model, we present the histogram of types of balanced triads 
  that emerge at the end of all simulations. 
  }
  \label{fig:hist-1}
\end{figure}

\begin{table}[t!]
    \centering
    \tiny
    \resizebox{0.60\linewidth}{!}{
    {\fontsize{9.5pt}{9.5pt}\selectfont
    \begin{tabularx}{0.6\textwidth}{c*{6}{>{\centering\arraybackslash}X}}
    \toprule
    &\multicolumn{2}{c}{\textbf{Rel.}}&\multicolumn{2}{c}{\textbf{Apprais.}}&\multicolumn{2}{c}{\textbf{Opin.}}\\
    \cmidrule{2-7}
    & \textbf{H} & \textbf{I} & \textbf{H} & \textbf{I} & \textbf{H} & \textbf{I}\\
    \midrule
\textbf{gpt-oss-120b}& {\subfloat{\includegraphics[width=0.95\linewidth]{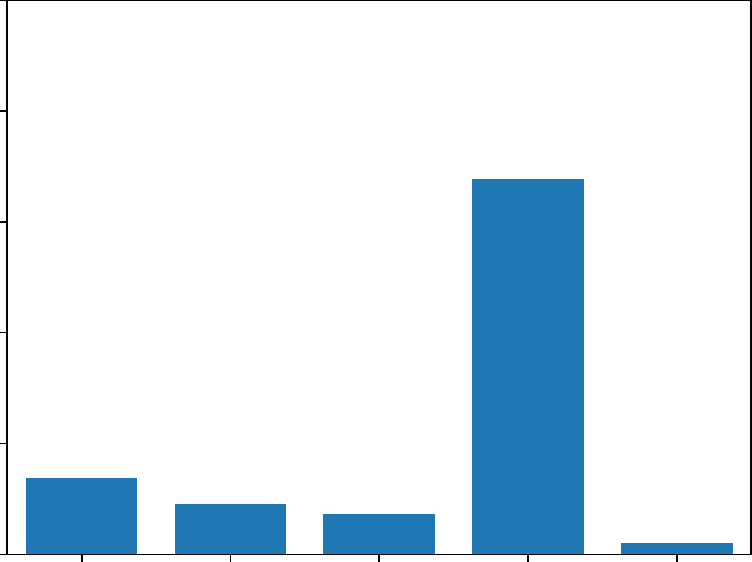}}} & {\subfloat{\includegraphics[width=0.95\linewidth]{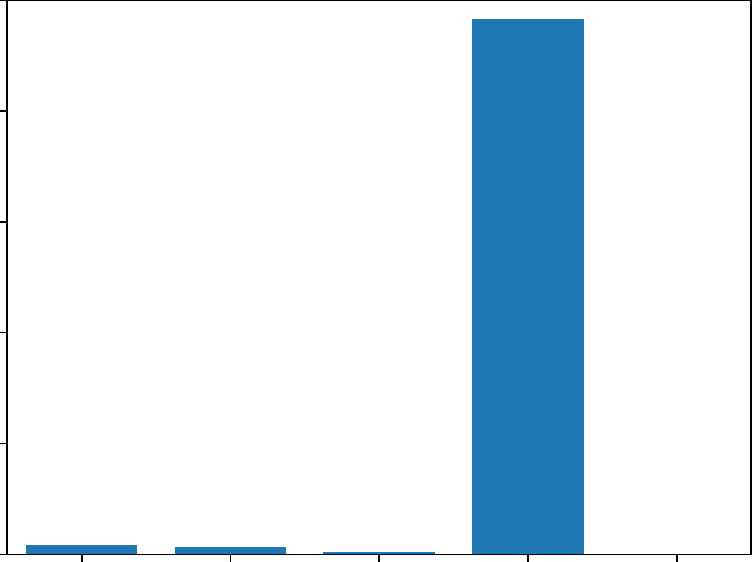}}}&{\subfloat{\includegraphics[width=0.95\linewidth]{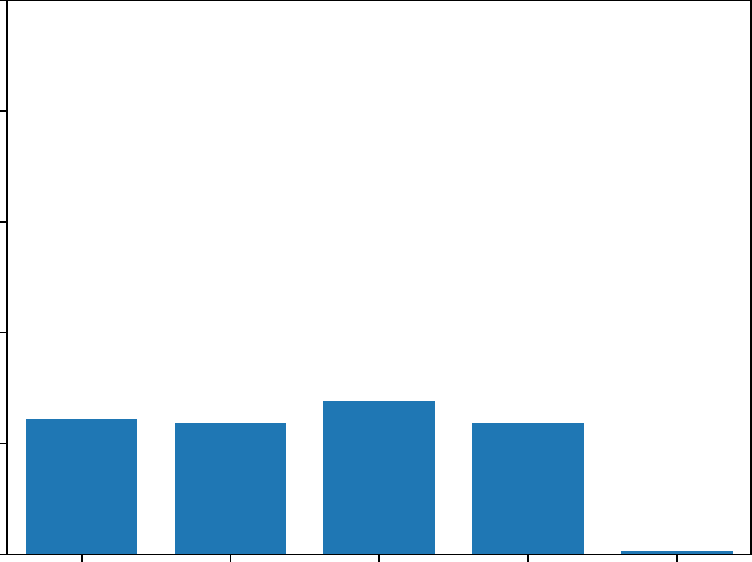}}}& {\subfloat{\includegraphics[width=0.95\linewidth]{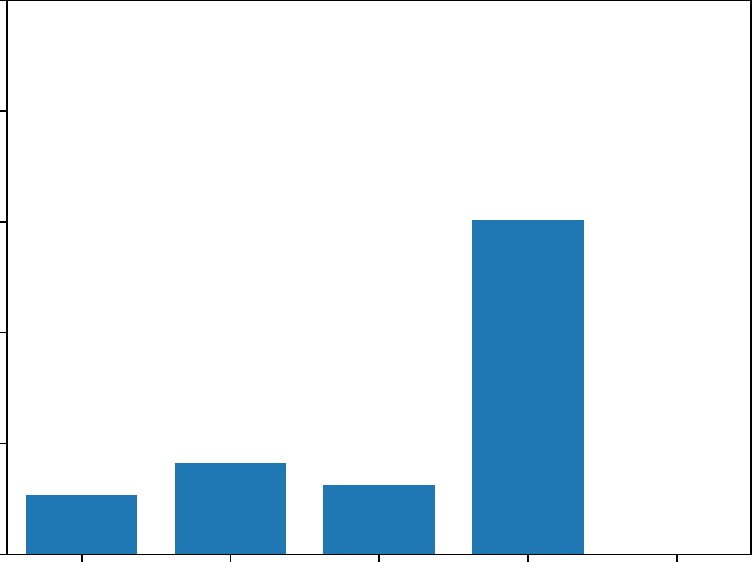}}}& {\subfloat{\includegraphics[width=0.95\linewidth]{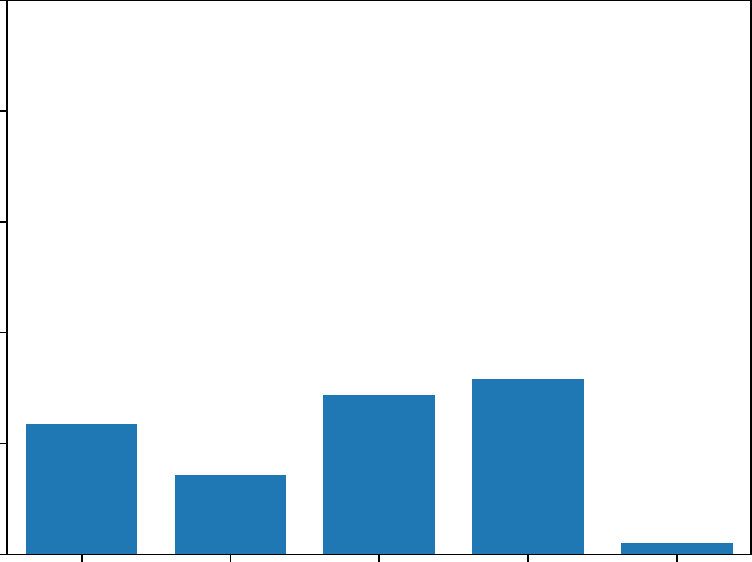}}}& {\subfloat{\includegraphics[width=0.95\linewidth]{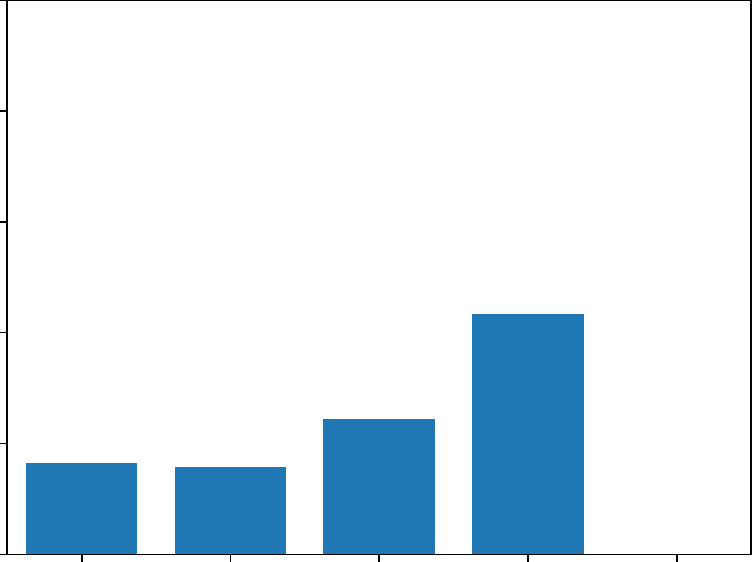}}}\\
\textbf{Llama~3~70B}& {\subfloat{\includegraphics[width=0.95\linewidth]{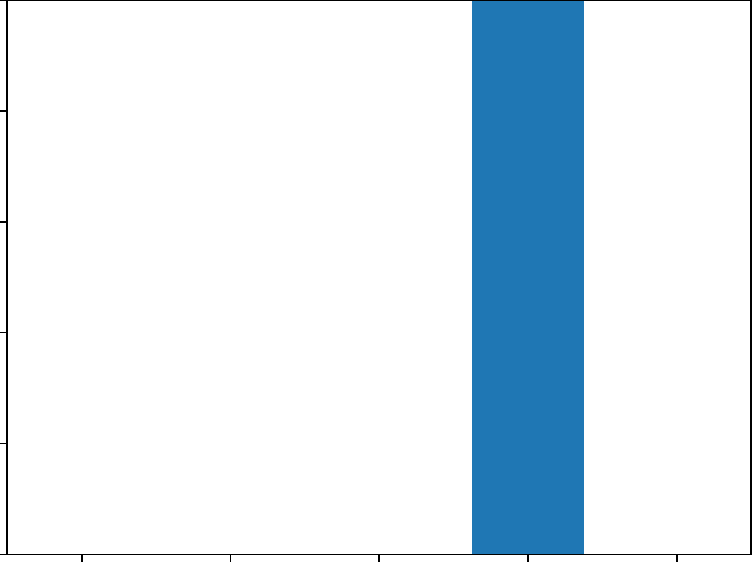}}}& {\subfloat{\includegraphics[width=0.95\linewidth]{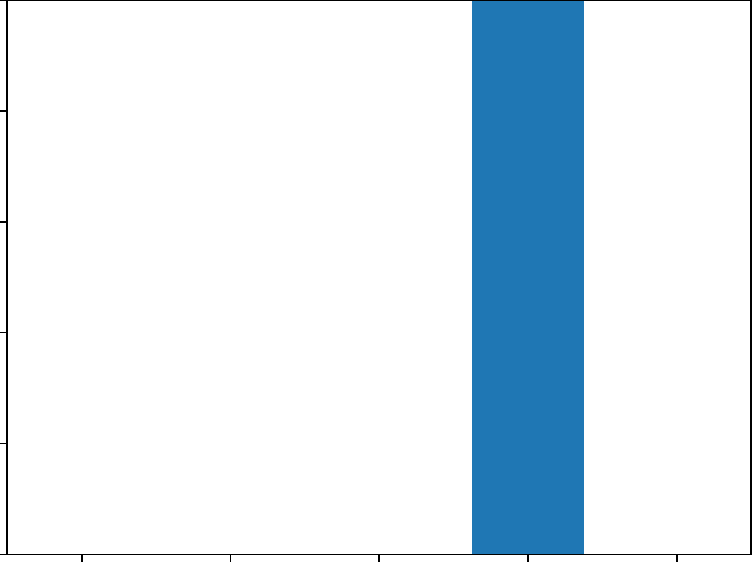}}}&{\subfloat{\includegraphics[width=0.95\linewidth]{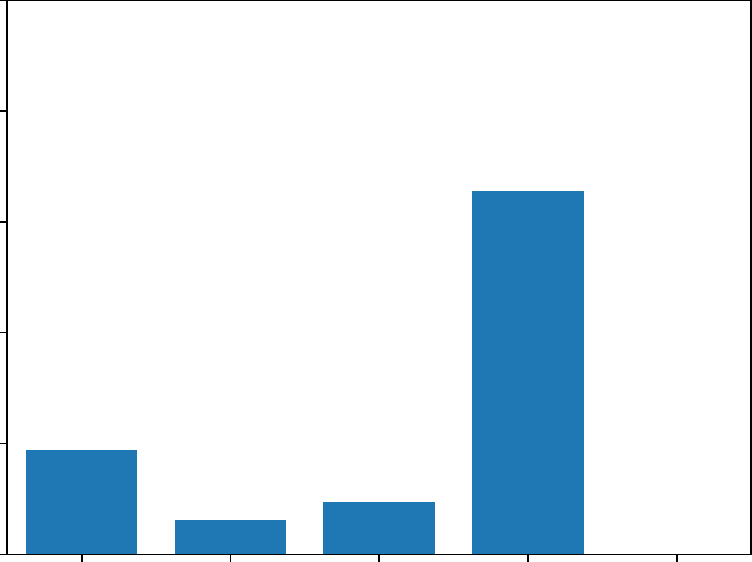}}}& {\subfloat{\includegraphics[width=0.95\linewidth]{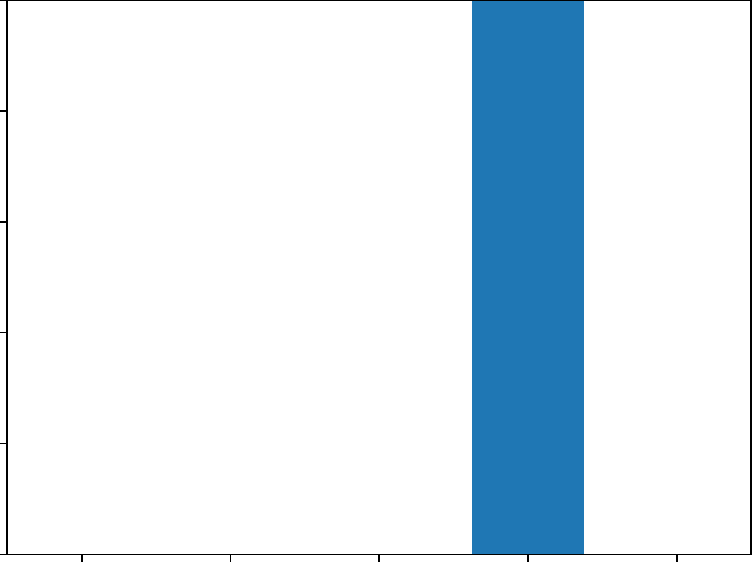}}}& {\subfloat{\includegraphics[width=0.95\linewidth]{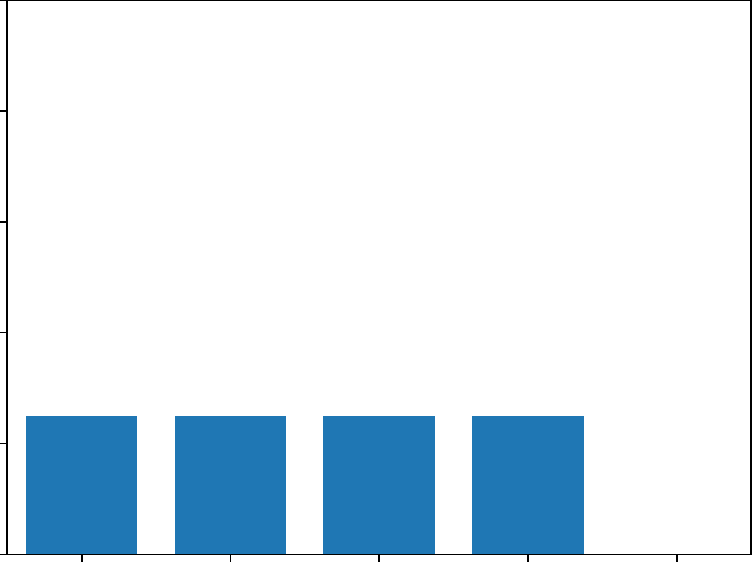}}}& {\subfloat{\includegraphics[width=0.95\linewidth]{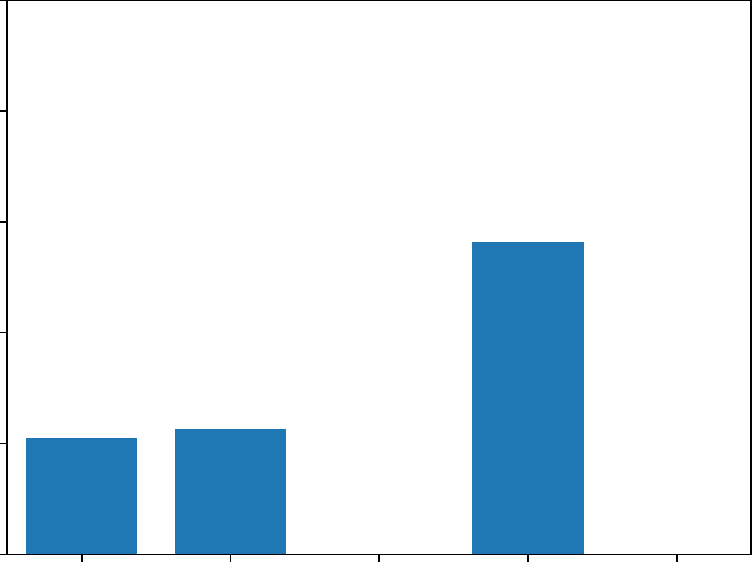}}}\\
\textbf{Llama~3~8B}& {\subfloat{\includegraphics[width=0.95\linewidth]{figs_hist/Hist_1_0.pdf}}}& {\subfloat{\includegraphics[width=0.95\linewidth]{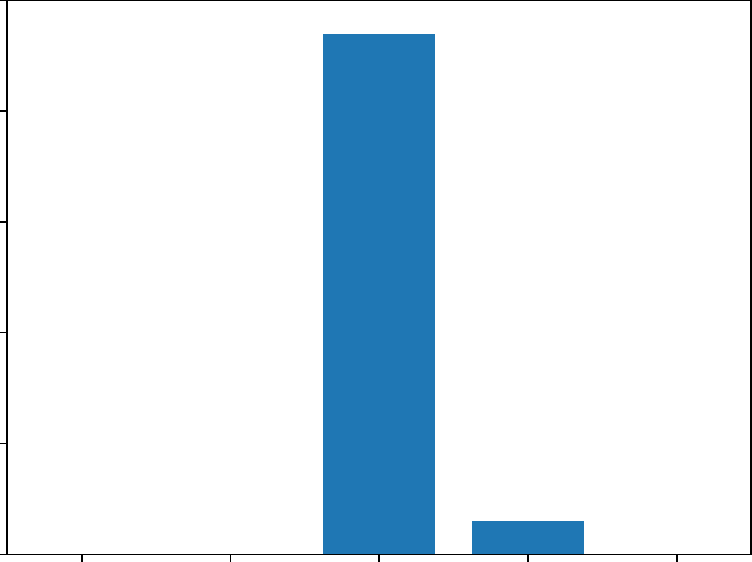}}}& {\subfloat{\includegraphics[width=0.95\linewidth]{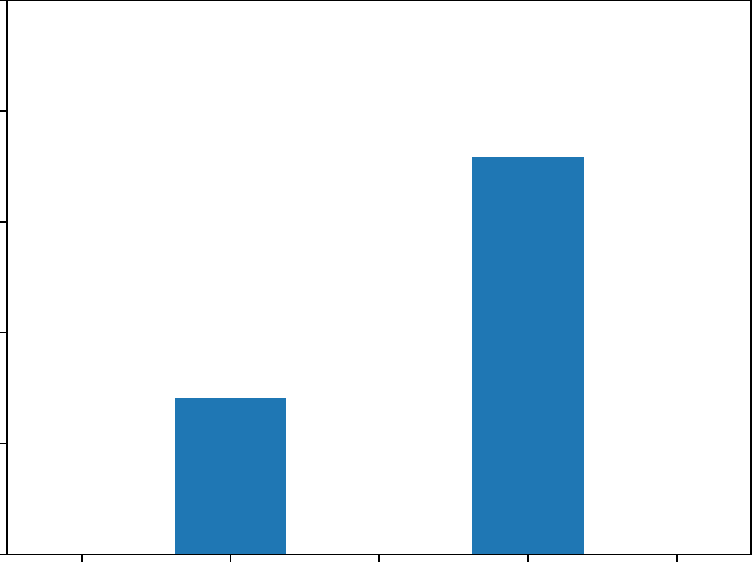}}}& {\subfloat{\includegraphics[width=0.95\linewidth]{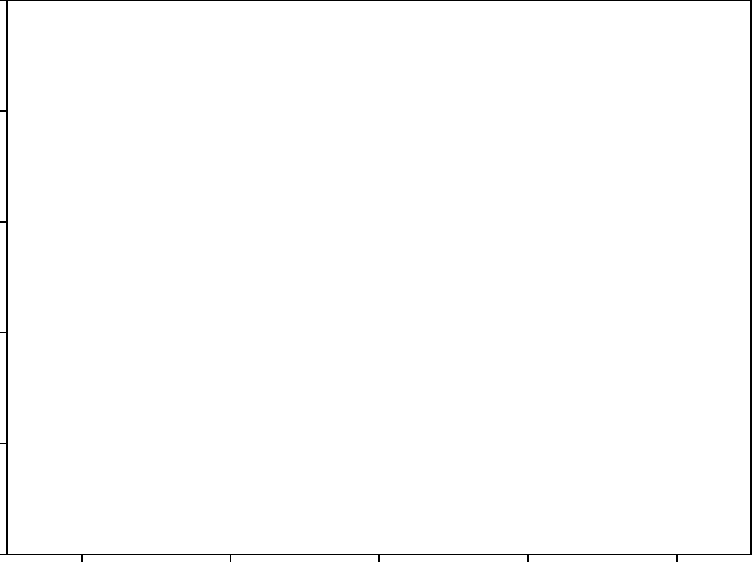}}}& {\subfloat{\includegraphics[width=0.95\linewidth]{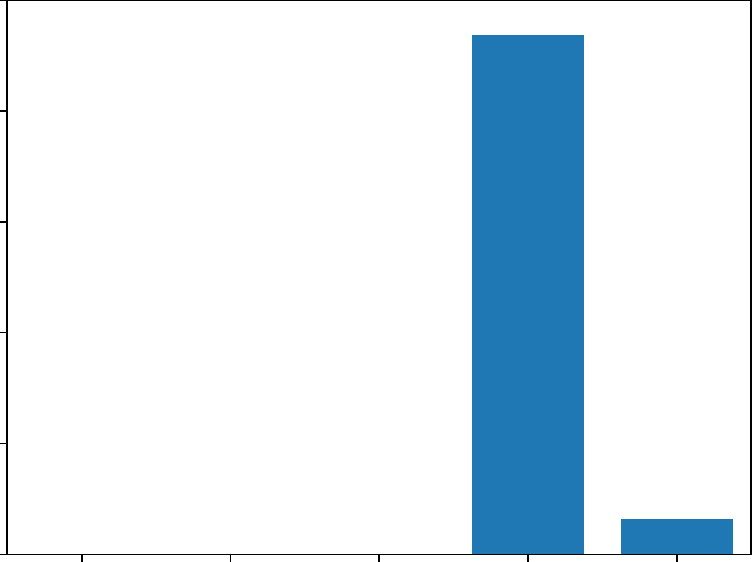}}}& {\subfloat{\includegraphics[width=0.95\linewidth]{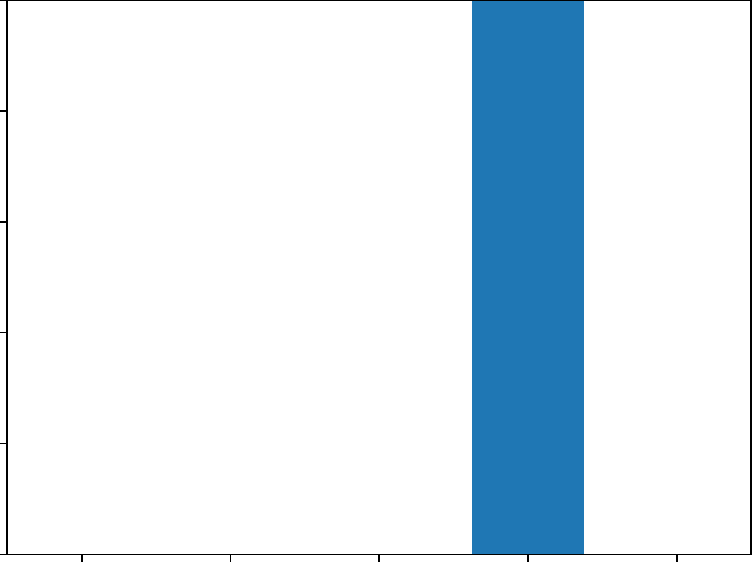}}}\\
\textbf{Mistral~7B} & {\subfloat{\includegraphics[width=0.95\linewidth]{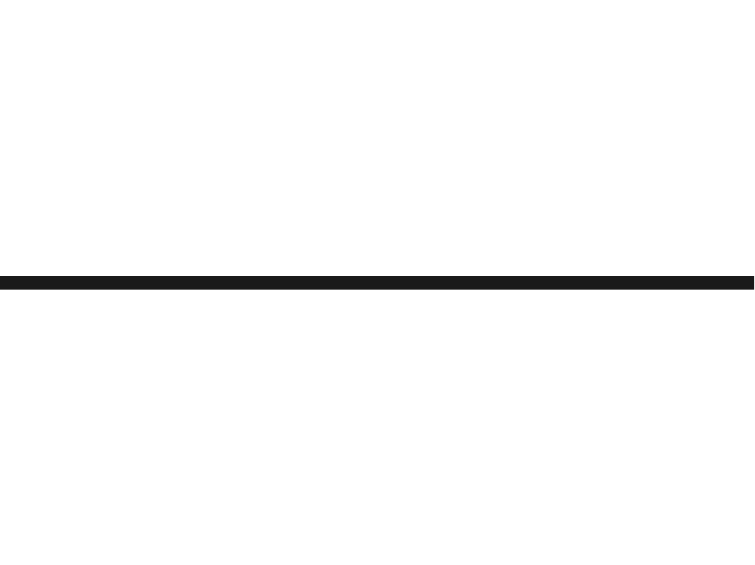}}}& {\subfloat{\includegraphics[width=0.95\linewidth]{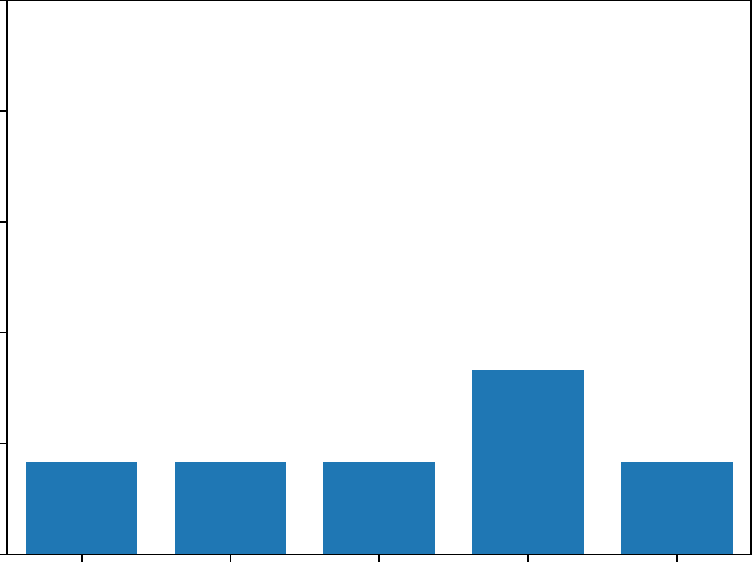}}}& {\subfloat{\includegraphics[width=0.95\linewidth]{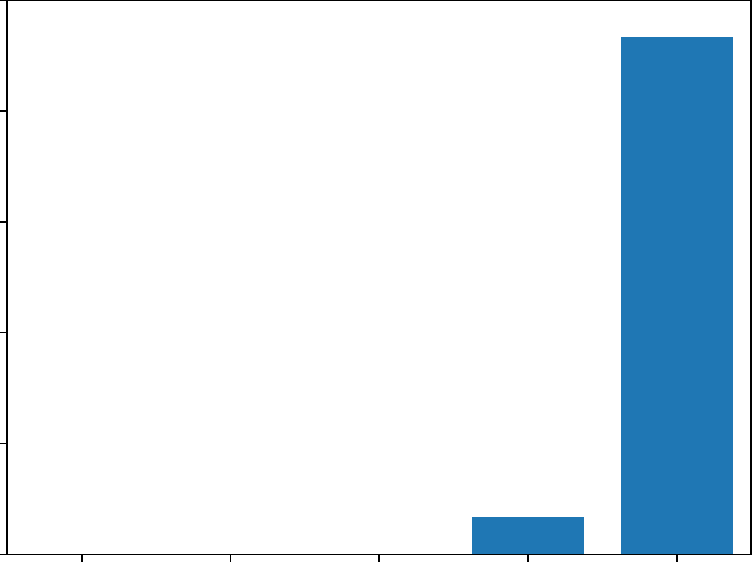}}}& {\subfloat{\includegraphics[width=0.95\linewidth]{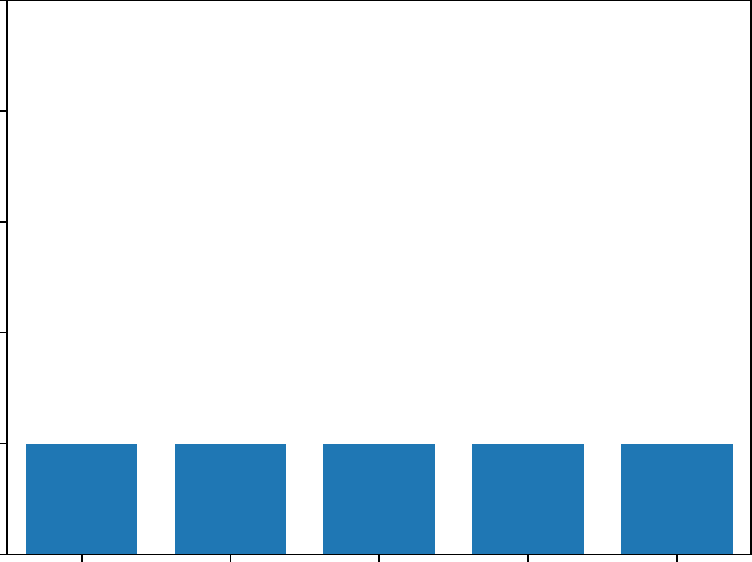}}}& {\subfloat{\includegraphics[width=0.95\linewidth]{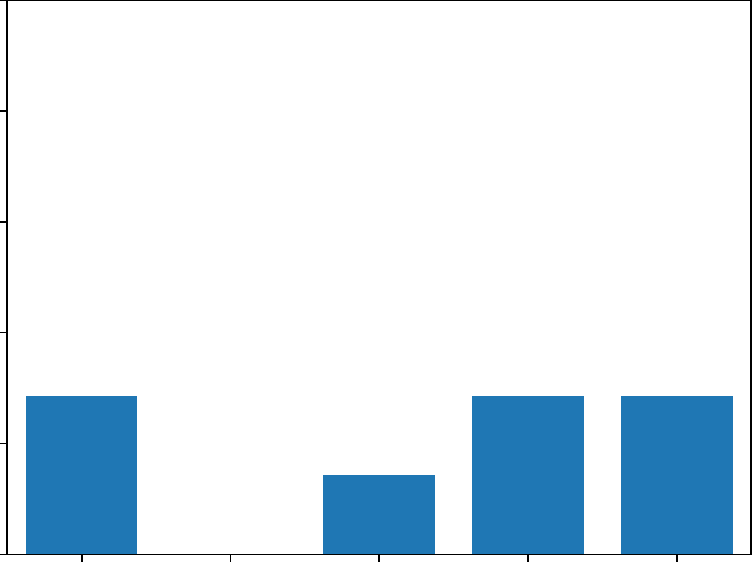}}}& {\subfloat{\includegraphics[width=0.95\linewidth]{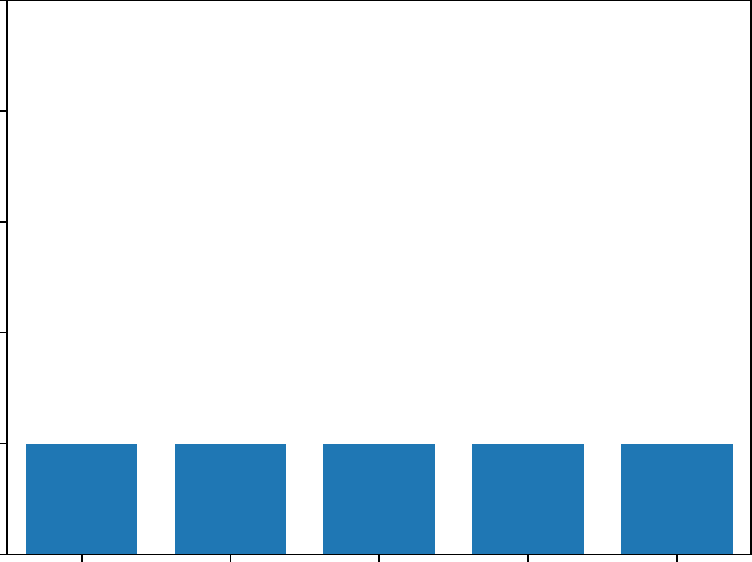}}}\\
\textbf{Gemma~3~4B}& {\subfloat{\includegraphics[width=0.95\linewidth]{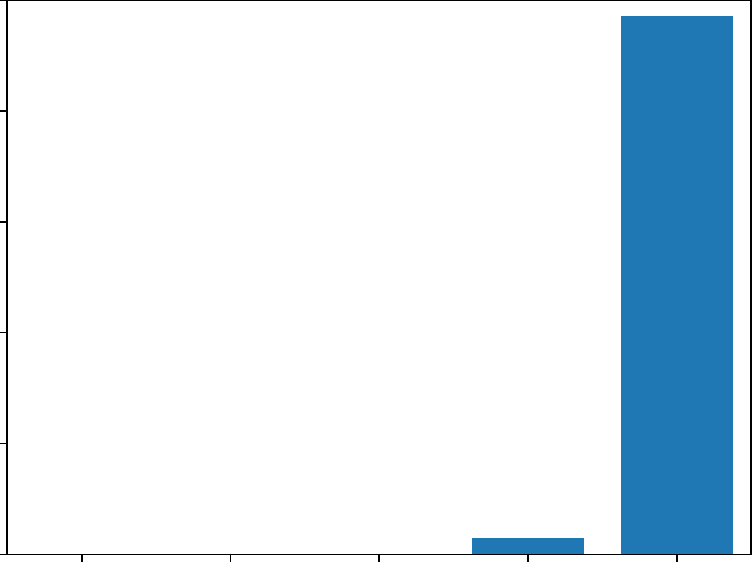}}} & {\subfloat{\includegraphics[width=0.95\linewidth]{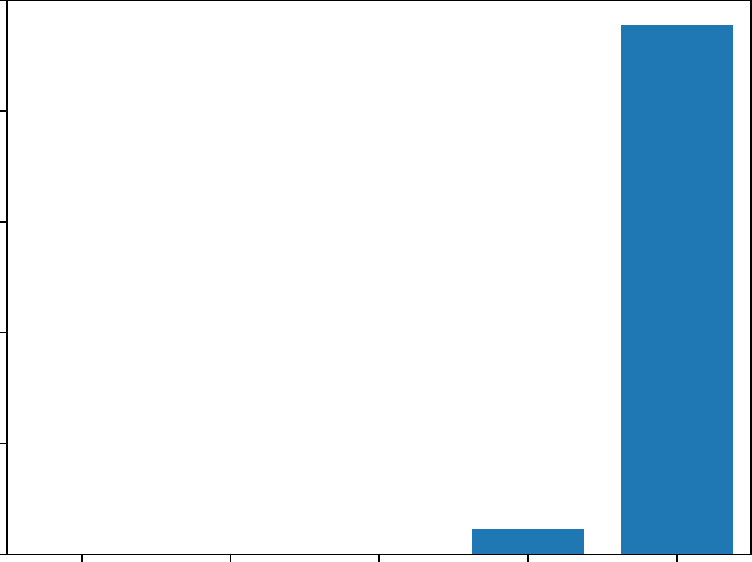}}}& {\subfloat{\includegraphics[width=0.95\linewidth]{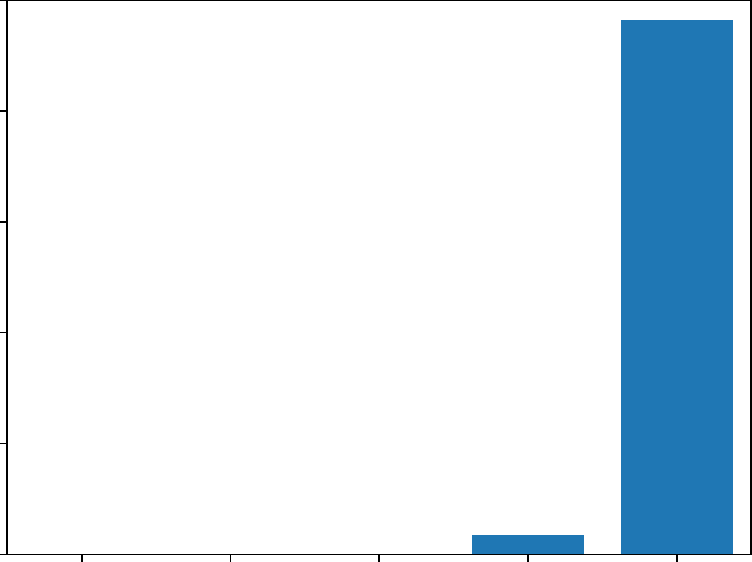}}}& {\subfloat{\includegraphics[width=0.95\linewidth]{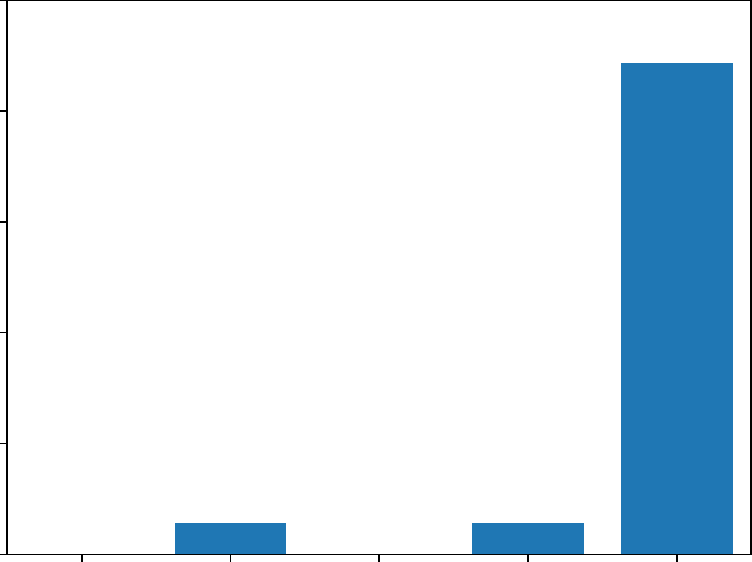}}}& {\subfloat{\includegraphics[width=0.95\linewidth]{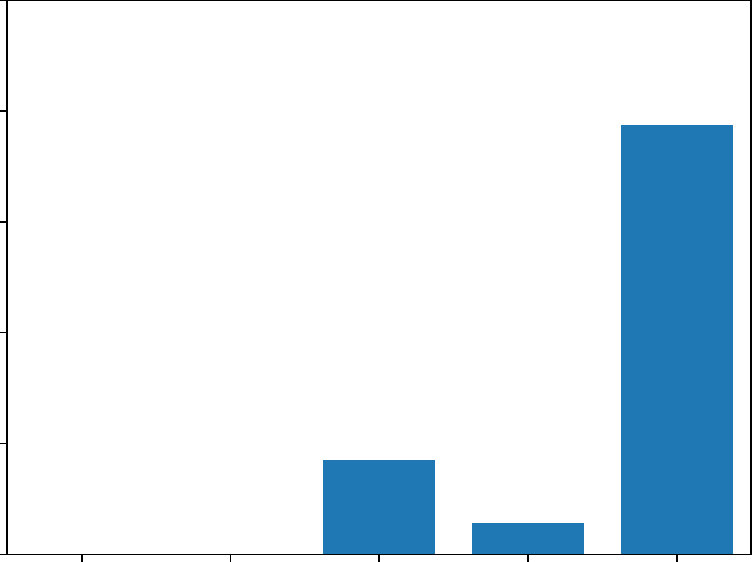}}}& {\subfloat{\includegraphics[width=0.95\linewidth]{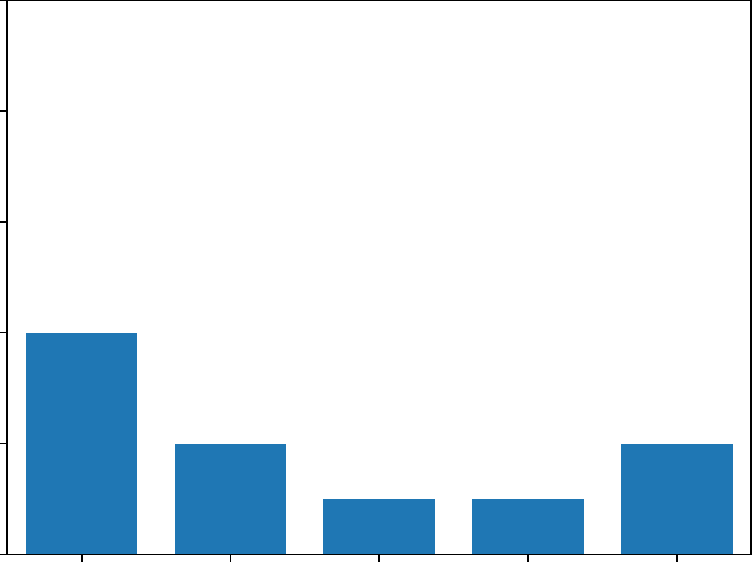}}}\\
\bottomrule
    \end{tabularx}
    
    }
    }
    \caption{
    \textbf{Diversity of balanced triads.}
    Each histogram has the x-axis  
    corresponding to the five balanced triads as in Fig.~\ref{fig:hist-1} and the y-axis going from $0\%$ to $100\%$ frequency. 
    The setting is similar to Table~\ref{tab:single-triad-freq}. 
    }
%
    \label{tab:hist-1-main}
\end{table}

We first study 
a population of three 
LLM agents. 
Additional information is in App.~\ref{app:more-sec3}.

\subsection{Achieving social balance}
\label{subsec:convBal}

We find that all LLM models achieve the well-established sociological phenomenon of social balance; see Tables~\ref{tab:single-triad-freq} and~\ref{tab:hist-1-main}, and Fig.~\ref{fig:hist-1}. 
%
Moreover, the \emph{type} of social balance, the \emph{frequency} at which it is achieved, and the \emph{diversity} 
of achieved 
balanced triads differ among models. Thus, 
\lightp{\emph{social balance can be used as an appropriate 
tool to 
characterize the social dynamics 
embedded in LLMs.}}

Regarding the \textbf{type} of social balance, gpt-oss-120b and Llama~3~70B achieve structural balance (no triad has all negative interactions) the most across types of 
interactions and update mechanisms, whereas the smaller 
Mistral~7B and Gemma~3~4B favor clustering balance (Tables~\ref{tab:single-triad-freq} and~\ref{tab:hist-1-main}, and Fig.~\ref{fig:hist-1}). 
Moreover, gpt-oss-120b and Llama~3 family lead to an overwhelming presence of all-positive triads (i.e., one faction), whereas Gemma~3~4B to all-negative ones (i.e., three factions) (Fig.~\ref{fig:hist-1}). 
Mistral~7B is more evenly distributed. 
Thus: \lightp{\emph{(i) largest models are likely to enforce the strongest concept of social balance, and (ii) strongly aligned models are biased to provide a more 
extreme positive/negative social outcome.}}\footnote{To the best of our knowledge, only Mistral~7B has not undergone strong safety alignment.} 

Regarding the \textbf{frequency} 
of achieving social balance,
gpt-oss-120b and Llama~3 family achieve social balance in the majority of 
simulations for most settings (boldface/underline in Table~\ref{tab:single-triad-freq}), 
whereas this is not true for 
other models 
(Table~\ref{tab:single-triad-freq}). 
%
%
%
%
%
%
%
%
We also find two surprising observations relating the frequency and \textbf{diversity} of achieved balanced triads. 
First, 
in the case of opinion interactions, Llama~3~8B 
achieves social balance more often than gpt-oss-120b and Llama~3~70B (Table~\ref{tab:single-triad-freq}); however, 
these two models
achieve more diverse sign configurations (Table~\ref{tab:hist-1-main}). Similarly, Mistral~7B has more diversity in settings where it achieves social balance less often. Surprisingly, gpt-oss-120b achieves both more diversity and frequency under appraisals than other models.
%
Thus: \lightp{\emph{(i) achieving social balance more often does not imply achieving it with more diversity of interactions and vice versa; 
(ii) model size is not a reliable predictor for achieving more diverse 
social outcomes.}}

One last observation about \textbf{diversity} of balanced triads: for a \emph{fixed} type of interaction (Table~\ref{tab:hist-1-main}), 
%
\lightp{\emph{the same update mechanism can (i) affect the diversity of balanced triads differently across models, and (ii) incite 
more positive interactions 
in one 
model 
and 
more negative ones 
in another.}}
%




\subsection{Stability of interactions}
\label{sec:stabl-interact}
%
%

Having shown that social dynamics of LLM agents 
can \emph{lead} to social balance, we study 
its \emph{stability}.

Our \emph{first} 
%
stability criterion 
%
%
analyzes 
how often 
an initially balanced triad 
does not change any of its interactions after ten iterations. 
The stability (decreasing) order is: Mistral~7B (stable 84.00\% of cases) $\succ$ Gemma~3~4B $\succ$ gpt-oss-120b $\succ$ Llama~3~70B $\succ$ Llama~3~8B (16.67\%).
Our \emph{second} 
stability criterion measures 
how often 
LLM agents do not change the sign of their interactions 
in the last five iterations, 
regardless of the final triad being balanced or not.
The stability (decreasing) order is: Mistral~7B (stable 99.78\% of cases) $\succ$ Llama~3~70B $\succ$ Gemma~3~4B $\succ$ gpt-oss-120b $\succ$ Llama~3~8B (49.95\%).

Our 
stability
results 
bring a new perspective on 
social dynamics: \lightp{\emph{(i) the model that achieves social balance more often 
is not necessarily the one with more stable interaction updates}};
and 
\lightp{\emph{(ii) LLMs are able to establish substantially stable social interactions despite their intrinsic stochastic nature.
}}
Fig.~\ref{fig:ex-conv} in App.~\ref{app:more-sec3} shows stability examples.



\subsection{The justification of interaction updates}
\label{sec:keyword-an}

\begin{table}[t!]
    \centering
    \def\arraystretch{1}
    
     \small
    \resizebox{0.6\linewidth}{!}{
    \begin{tabularx}{0.6\linewidth}{c
    *{6}{>{\centering\arraybackslash}X}{c}}
    \toprule
    & \multicolumn{2}{c}{\textbf{Rel.}} & \multicolumn{2}{c}{\textbf{Apprais.}} & \multicolumn{2}{c}{\textbf{Opin.}} \\
    \cmidrule{2-7}
    & \textbf{H} & \textbf{I} & \textbf{H} & \textbf{I} & \textbf{H} & \textbf{I} \\
    \midrule
    \textbf{gpt-oss-120b-J}
    & 0.03 & $\times$ & 0.71 & 0.07 & 0.76 & 0.01\\
    \textbf{gpt-oss-120b-R} 
    &  0.66 & 0.15 & 5.14 & 0.55 & 2.96 & 0.12\\
    \textbf{Llama~3~70B} 
    & $\times$ & $\times$ & 13.80 & 5.28 & 6.28 & 1.16\\
    \textbf{Llama~3~8B}  
    & $\times$ & $\times$ & 4.10 & $\times$ & 1.90 & 0.01\\   
    \textbf{Gemma~3~4B}  
    & $\times$ & $\times$ & 4.60 & 0.23 & 3.06 & 0.94\\
    \bottomrule
    \end{tabularx}}
    %
    %
        \caption{\textbf{Frequency (percentage \%) of responses with the keyword ``dissonance''.} gpt-oss-120b: response justification (J) and reasoning (R). ``$\times$'' indicates $0.00$.
    } 
    \label{tab:keyword-cogdis-main}
\end{table} 

For the sake of interpretability, we focus on finding evidence of 
LLM agents 
\emph{expressing} an intention to achieve social balance. 
%
%
%
Particularly, we look 
for relevant keywords 
in every justification given by every agent when updating its social interactions. 
%
%
%
In this section we only consider the keyword ``dissonance'' which is argued to occur when Heider's rules are violated (Section~\ref{sec:arXiv_intro}); other keywords are analyzed in App.~\ref{app:more-sec3}.
Keyword 
``dissonance'' is found in Llama~3 family and Gemma~3~4B under appraisals and opinions; see Table~\ref{tab:keyword-cogdis-main}. 
Coincidentally, appraisals and opinions lead to a 
larger diversity of 
balanced triads than relationships 
(Table~\ref{tab:hist-1-main}). 
Moreover, this keyword appears most often 
in Llama~3~70B 
in the same setting that leads to social balance with $100\%$ frequency (Table~\ref{tab:single-triad-freq}). 
%
%
Remarkably, 
``dissonance'' 
is used to justify interaction updates that \emph{explicitly resolve} a violation of Heider's rules (enforcing social balance), though in Gemma~3~4B the opposite is also often observed; see App.~\ref{sub:ka}. 
%
Interestingly, gpt-oss-120b uses a mathematical approach that calculates cycle signs in order to enforce Heider's rules---probably due to being a reasoning model (App.~\ref{sub:ka}). 
%
%
%

Thus, 
\lightp{\emph{LLMs can show 
knowledge of sociopsychological 
processes that lead to social balance, 
and this 
could benefit both diversity and frequency of balanced triads.}}

%% file: sec/multiple.tex
All LLM models achieve social balance for 
populations of 20 and 120 triads, i.e., of six and ten agents, respectively; 
see Table~\ref{tab:multiple-pop-type} in App.~\ref{app:tabl-multi}.
%
Social balance is achieved if \emph{every} single triad is balanced.
%

Two notable differences exist compared to the single triad case.
First, 
the smallest models achieve social balance more often than the largest ones across most settings, even when the larger ones do not 
(Table~\ref{tab:multiple-pop-type}). 
Second, 
only gpt-oss-120b and Llama~3~70B achieve all-positive interactions, and only in one setting. 
Llama~3~70B now achieves clustering balance 
(Table~\ref{tab:multiple-pop-type}). 

Thus, our results indicate that,  
\lightp{\emph{in the more complex social scenario 
of 
populations with multiple triads, 
the LLM models display social dynamics  
different from the single triad case when updating interactions.}}
Moreover, \lightp{\emph{unlike the single triad case, a larger model does not necessarily achieve social balance more often and in a stronger type.}}

%
Llama~3~70B and gpt-oss-120b only report ``dissonance'', and in a tiny amount:
no more than 
$0.17\%$ 
per setting in Llama~3~70B---compared to 
the $13.80\%$ for a single triad (Table~\ref{tab:keyword-cogdis-main})---and considerably less 
in gpt-oss-120b; see Table~\ref{tab:keyword-dis-mult} in App.~\ref{app:tabl-multi}. 
%
%
%
%
\lightp{\emph{This supports the existence of different social dynamics when LLMs belong to larger populations.}}
Other keywords are analyzed in App.~\ref{app:tabl-multi}.

\section{Further Discussion}
\label{sec:disc-main}

We provide a detailed example of how our results 
inform an agentic application in social media 
in App.~\ref{sec:disc}. 
Our work's motivation is further explained in App.~\ref{app:furthmotiv-0}. Our choice of LLM models, update of interaction signs, and display of social information are further discussed in App.~\ref{app:further-motiv}.

\section{Conclusion}
We show that LLMs can achieve social balance after the continuous updating of their social interactions. We characterize how the frequency, diversity, and stability of balanced triads vary across models. We also find that
changes in population size can induce different social dynamics. As future work, we propose
studying how achieving social balance is affected by LLMs having memory of past social interactions

\section*{Acknowledgments}
We thank the VMware Research Group. We also thank the people at VMware involved in the deployment of large
language models for providing us with adequate computational resources to run the models and to all those who
provided us with any information regarding the use and the specifications of the platform used in this study. Finally, we thank Jessica C. for some improvements on the reading of the paper.

%% file: app/further-motiv_0.tex
\paragraph{The focus on social balance.}Our paper aims to provide an answer to the \emph{general question} of understanding how LLMs process signed 
interactions in a multi-agent setting. 
The fact that social balance \emph{emerges} from the evolution of interactions is remarkable---when running our experiments there was \emph{no prior guarantee or knowledge} that social balance would emerge. Therefore, given that (i) social balance is a well-established concept in the sociological literature and that (ii) it actually emerges in multi-agent LLM systems, we decided to place it as the \emph{scope} of our paper. 

Even within this scope, our work shows that LLM models display a variety of behaviors on \emph{how} they achieve social balance (Sections~\ref{sec:triad} and~\ref{sec:mult}), thus showing that social balance is a framework \emph{rich enough} to capture diverse social dynamics embedded in LLMs.

We also remark that 
previous works have studied LLMs in social scenarios such as opinions dynamics in multi-agent systems (App.~\ref{sec:arXiv_related}
). Our work \emph{expands} this literature to the dynamics of sociologically relevant \emph{signed} 
interactions across LLMs.

\paragraph{A connection with size complexity.} It 
is known that 
some LLMs' abilities can improve and new ones can emerge as LLMs grow in size~\citep{wei-2022-emergent}. 
Our contributions show this is not necessarily true for 
collective emergent properties such as 
achieving social balance.

\paragraph{A connection with alignment.} 
Our work 
contributes to the characterization of alignment in LLMs~\citep{liang-2023-holistic}: 
how \emph{effectively} an LLM model achieves social balance could be used as a metric for how alignment \emph{affects} social dynamics. 
For example, our results show that LLMs that underwent a strong alignment procedure has either an extremely positive or negative outlook in their interactions (Section~\ref{sec:triad}).  
%
Note that in our work 
LLMs are 
only affected by ``built-in'' biases, i.e., the \emph{default} ones coming from alignment, since there is no 
assigned persona~\citep{aher-2023-llmsimulatehuman} 
that could
introduce 
additional biases~\citep{salewski-2023-incontext,chuang-2024-simulating}.

\paragraph{Our focus on LLM-to-LLM interactions.} 
Our work only considers LLM-to-LLM interactions because it is a \textbf{controlled environment to analyze LLM multi-agent behavior}. This setting is also relevant for two reasons: 
\begin{enumerate}
    \item We believe it will be \textbf{more common} to find LLM-to-LLM interactions in social media in the near future, e.g., more users may have LLM agents posting and extracting information for them.
    \item In our work, \textbf{the LLM does not know whether it is interacting with a human or another LLM}: nowhere in the prompts given to the LLMs it is indicated the nature of the ``individuals'' the LLM is interacting with (Apps.~\ref{subs:prompt-det-Llama} and~\ref{subs:prompt-det-Mistral}). In other words, the LLM is \emph{agnostic} to the identity of the agents that interact with it. Thus, when an LLM explicitly changes its interactions in order to satisfy Heider’s rules (e.g., by reducing cognitive dissonances; Section~\ref{sec:keyword-an}), \textbf{the same behavior is expected to occur even if humans were involved as interacting agents}.
\end{enumerate}

%% file: app/rel_works.tex
\paragraph{Dynamics of LLM interactions in multi-agent settings.} To the best of our knowledge, no prior work has focused on how LLMs update different types of \emph{signed} (positive or negative) interactions (i) without taking a specific persona, (ii) by only observing the current state of such interactions, and (iii) under the framework of social balance. 
The works~\citep{chuang-2024-simulating,chuang-2024-wisdom} focus on how opinions spread and change among LLMs role-playing different persona, and the effect caused by injecting external biases and by the degree of impersonation. The work~\citep{zhou2024sotopia} studies the display of social intelligence in role-playing LLMs, and, among different variables, explicitly considers the effect of family, friend, romantic, acquaintance,
and stranger relationships among the agents. 
The work~\citep{cisnerosvelarde2024princopindynLLM} takes the LLM as is, and studies the conflict among biases present during opinion exchange. Such work considers opinions that display positive and 
negative connotations.
%
%
%
%
%
%
%
%

\paragraph{Dynamic modeling of signed interactions.} Several 
works in mathematical sociology 
study 
how agents 
process 
signed interactions. 
One 
line of research studies how established positive and negative ties
among agents affect the diffusion of opinions; e.g., social balance 
is known 
to lead to polarization of opinions~\citep{guodongalta-DynSignNet-2019,cisneros-velarde-2021-polarization}. 
Another 
line of research studies models that evolve signed 
interactions among agents toward structural balance~\citep{kulakowski2005heider,marvelkleinberg-2011-ctmodelstbal,traagvandooren-2013-DynModExpSocBal,mei-2019-DynSocBal,cisneros-StBviaGradientFlow-2021} and clustering balance~\citep{vanderijt-2011-micromacro,cisnerosvelarde-2020-SignFormClustering}. 
Generally, all these works assume that the mathematical equations for the update of interactions
are tractable and sociologically plausible,   
which is difficult to do in non-linear stochastic models with billions of parameters such as transformer-based LLMs \citep{vaswani-2017-attention}.

\paragraph{Social balance and signed networks.} 
Structural balance and clustering balance were formulated in~\citep{heider-1946-AttCogOrg} and~\citep{davis-1967-cluststructbalgraphs}, respectively. Textbook treatments of social balance can be found in~\citep{Kleinberg2010NetCr,Friedkin1998StrBBook},
and a comprehensive survey on both theoretical and empirical works in~\citep{Zheng2015-nk}. 
%
%
The presence of social balance has been used as a metric of social network analysis for signed networks
~\citep{Camacho2020FourDim}. Indeed, a positive/negative link between individuals not only could indicate that one agent thinks of the other as a friend/enemy, but also as someone considered to have high/low prestige or trustworthiness~\citep{guha2004TrustDistrust}. 
%
%
%
In applications with large datasets of interactions across members of a social network, both the signs of such interactions  
~\citep{hassan-etal-2012-extracting,yang2012FriendFrenemy} and the 
extraction of the signed network structure itself
~\citep{junjiehuang2021SDGNN,huang2022POLE} need to be inferred
from texts; for a historical survey we refer to~\citep{Tang2016Survey}. 
As application examples, 
the presence of social balance has been used to analyze international relationships
~\citep{Gallo2024-testingSB,Askarisichani2020-ut,moore1978internapplheider,Maoz2007EnemyEnemy,doreian2015StbakanceSignedRel}, online social networks~\citep{Gallo2024-testingSB,huang2022POLE,SzellLambiotte-2010-MuiltOrgLScaleSocOn,leskovec-2010-SignedNetSocMedia}, consumer behavior~\citep{Fink2009OffFieldBehavior,Woodside2004ConsBraBalance}, 
and animal social behavior~\citep{Ilany2013StrBalanceSocAnimal}.

\paragraph{LLMs as Economic Agents.} Although we do not study strategic decision making of LLMs, this is a growing field related to 
multi-agent LLMs in sociologically relevant scenarios. 
For example,
%
\citep{li-etal-2024-econagent} uses agent-based modeling with LLMs to simulate macroeconomic phenomena; \citep{xie2024can} studies trust behavior in game theoretic settings; and \citep{zhao2024competeai} studies LLMs as competing businesses. For more related works, we refer to the survey~\citep{llmsum-2024-surv}.



%% file: app/more-info-sec3.tex
The information that complements our treatment from  Section~\ref{sec:triad} is divided according to subsections with their same equivalent name in the main paper.

\subsection{Achieving social balance}

Regarding Table~\ref{tab:single-triad-freq}, the shown percentages are out of $640$ simulations: $10$ simulations for each of the $2^6$ initial conditions (Section~\ref{sec:prob-sett}).

Regarding Fig.~\ref{fig:hist-1}, the percentages are out of a total of $3840$ ($3200$) simulations for Llama~3, gpt-oss, and Gemma~3~4B (Mistral~7B): $6$ ($5$) settings of $640$ simulations each one.

We remark that only Mistral~7B refuses to provide an answer to our request under relationships with homophily, e.g., by stating that a new relationship would be ``uncertain'' or impossible to determine given the information provided (in the prompt).

\subsection{Stability of interactions}
\label{aoo:stabil}

\begin{figure}[t]
\captionsetup[subfigure]{labelformat=simple} 
\renewcommand\thesubfigure{}
    
    \centering
    %
\subfloat[\centering  Stable]{\includegraphics[width=0.22\columnwidth]{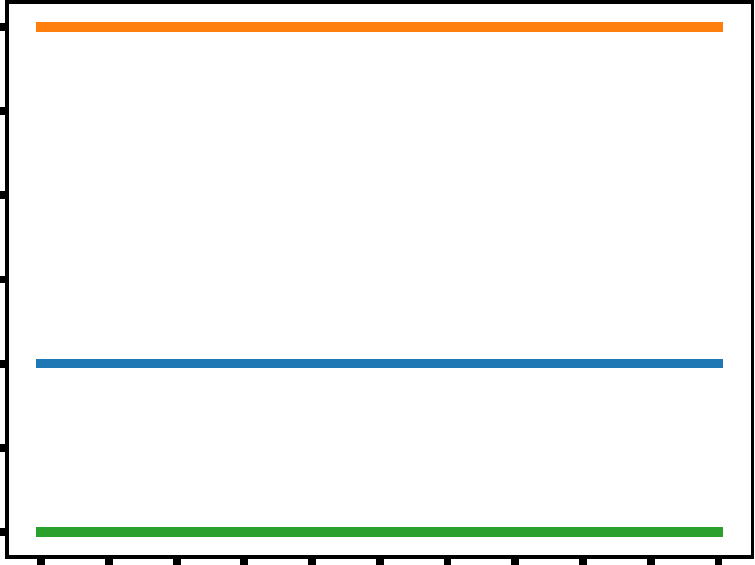}}
\hspace{2pt}
\subfloat[\centering  Unstable]{\includegraphics[width=0.22\columnwidth]{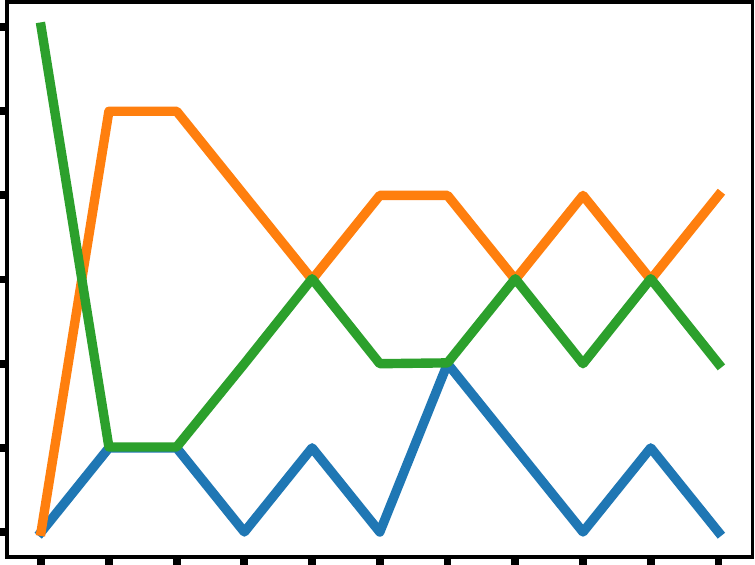}}
\hspace{3pt}
\subfloat[]{\includegraphics{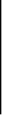}}
\hspace{3pt}
\subfloat[\centering  Stable]{\includegraphics[width=0.22\columnwidth]{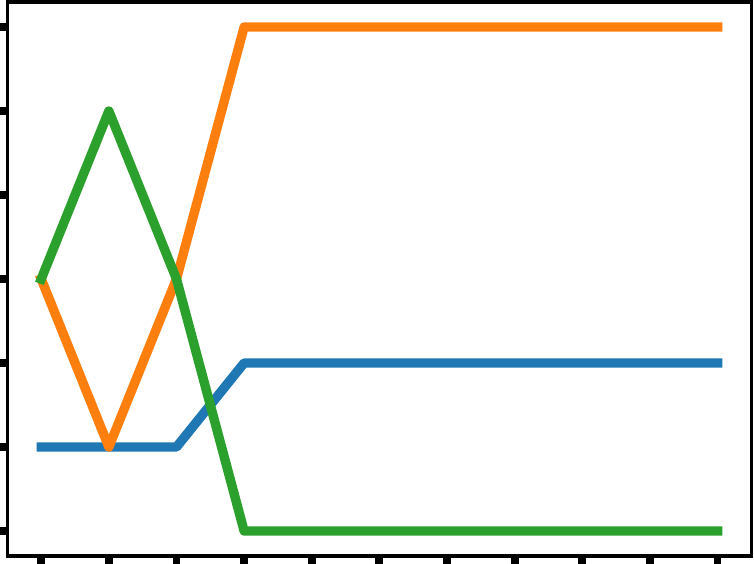}}
\hspace{2pt}
\subfloat[\centering Unstable]{\includegraphics[width=0.22\columnwidth]{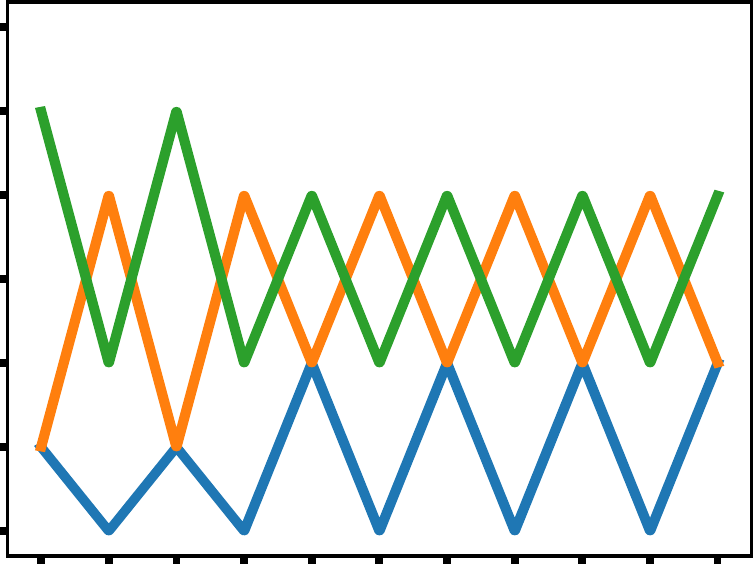}}
\caption{\textbf{Examples: 
stability of interactions for gpt-oss-120b (left) and Llama~3~70B (right) according to the second criterion of Section~\ref{sec:stabl-interact}.
} Changes on the number of positive cycles (blue curve), and positive (orange) and negative (green) interactions across ten iterations (x-axis, starting at $t=0$) and taking values from $0$ to $6$ (y-axis). 
%
%
%
%
Social balance is achieved at the tenth iteration in all cases.
}
    \label{fig:ex-conv}
    \end{figure}

Examples of the evolution of interactions and their stability is found in Fig.~\ref{fig:ex-conv}. We point out that a \emph{positive cycle} occurs when the product of the interaction signs, in a clockwise or anticlockwise direction, is positive; see App.~\ref{sec:prob-sett-exp}.

We now the discuss the number of simulations that were considered in our stability analyses. For our \emph{first} stability criterion, the total number of simulations is $300$ for all models except Mistral~7B: 
$10$ simulations for each of the $5$ types of balanced triads %
across the $6$ types of interactions and update mechanisms. For Mistral~7B it is $250$ because we 
only consider $5$ settings (Section~\ref{subsec:convBal}).
For our \emph{second} stability criterion, the total number of simulations is $3840$ for all models except Mistral~7B: $10$ simulations for each of the $2^6$ possible sign distributions over a triad across the $6$ types of interactions and update mechanisms. For Mistral~7B it is $3200$ 
because we only consider five settings.

According to the \emph{first} stability criterion, we find that gpt-oss-120b is stable on $49.00\%$ of cases, Llama~3~70B on $46.67\%$, Llama~3~8B on $16.67\%$, Mistral~7B on $84.00\%$, and Gemma~3~4B on $53.33\%$.
According to the \emph{second} stability criterion, we find that gpt-oss-120b is stable on $66.25\%$ of cases, 
Llama~3~70B on $92.19\%$, Llama~3~8B on $49.95\%$, Mistral~7B on $99.78\%$, and Gemma~3~4B on $85.31\%$. We remark that this second stability criterion is a conservative one since it looks for no social change during the \emph{last half} of all interactions.

\subsection{The justification of interaction updates}

We consider the obvious keywords ``structural balance'', ``clustering balance'', and ``social balance''. 
Surprisingly, \emph{no} LLM model mentions ``clustering balance'', despite achieving this social structure. 
gpt-oss-120b uses the term ``social balance'' when updating its interactions, in both its responses and reasoning; see Table~\ref{tab:keyword-socbal}. 
gpt-oss-120b takes a particular mathematical approach to detect Heider's rules' violations by analyzing positive cycles
\footnote{See App.~\ref{aoo:stabil} for the definition of positive cycles.}---probably due to being a reasoning model; e.g., see App.~\ref{sub:ka}.
%
Llama~3~70B only mentions ``social balance'' in the negligible amount 
of 3 out of all the 230400 responses (i.e., $0.001\%$); 
e.g., see App.~\ref{sub:ka} for examples. 
No other LLM mentions ``social balance''. Finally, only gpt-oss-120b mentions ``structural balance'' in smaller proportions; see Table~\ref{tab:keyword-socbal}. 

Two opposites results occur regarding 
``social balance'' and ``dissonance'': the former is more frequent in gpt-oss-120b than the LLama~3 family, whereas the opposite is for the latter (Tables~\ref{tab:keyword-cogdis-main} and~\ref{tab:keyword-socbal}).

\begin{table*}[t!]
    \centering
    \def\arraystretch{1}
    
     \small
    \begin{tabularx}{0.7\textwidth}{c
    *{5}{>{\centering\arraybackslash}X}{c} *{7}{>{\centering\arraybackslash}X}{c}
    }
    \toprule
     & \multicolumn{6}{c}{\textbf{``\texttt{social balance}''}} & \multicolumn{6}{|c}{\textbf{``\texttt{structural balance}''}}\\ 
    \cmidrule{2-13}
    & \multicolumn{2}{c}{\textbf{Rel.}} & \multicolumn{2}{c}{\textbf{Apprais.}} & \multicolumn{2}{c}{\textbf{Opin.}} &
    \multicolumn{2}{c}{\textbf{Rel.}} & \multicolumn{2}{c}{\textbf{Apprais.}} & \multicolumn{2}{c}{\textbf{Opin.}} \\
    \cmidrule{2-13}
    & \textbf{H} & \textbf{I} & \textbf{H} & \textbf{I} & \textbf{H} & \textbf{I} & \textbf{H} & \textbf{I} & \textbf{H} & \textbf{I} & \textbf{H} & \textbf{I} \\
    \midrule
    \textbf{gpt-oss-120b-J} & 5.44 & 3.10 & 0.38 & 1.18 & 0.85 & 0.50
    & 1.16 & 1.98 & 0.14 & 0.45 & 0.10 & 0.15\\
    \textbf{gpt-oss-120b-R} 
    & 3.07 & 2.98 & 0.75 & 1.46 & 0.30 & 0.36 
    & 1.65 & 3.16 & 0.21 & 0.71 & 0.27 & 0.28
    \\
    \bottomrule
    \end{tabularx}
    %
    %
        \caption{\textbf{Frequency (\%) at which the keywords ``social balance'' and ``structural balance'' appear for gpt-oss-10b response justification (J) and reasoning (R).} Each percentage is with respect to a total of $38400$ responses (there are 
    $10$ simulations for each of the $2^6$ possible sign distributions over a triad wherein $3$ agents update $2$ interactions across $10$ iterations). 
    } 
    \label{tab:keyword-socbal}
\end{table*}

\section{Additional Information for Section~\ref{sec:mult}}
\label{app:tabl-multi}

Table~\ref{tab:multiple-pop-type} and Table~\ref{tab:keyword-dis-mult} present diverse results for multiple triads population. 

We also find that  
no model reports ``clustering balance''. Only gpt-oss-120b reports ``structural balance'' and ``social balance'' at low frequency; see Tables~\ref{tab:keyword-gpt-mult-6} and~\ref{tab:keyword-gpt-mult-10}. Despite mentioning such keywords, gpt-oss-120b achieves social balance less often than all other models.

\begin{table*}[t!]
    \centering
    \def\arraystretch{1}
    
     \small
    \begin{tabularx}{0.7\textwidth}{@{}c
    *{5}{>{\centering\arraybackslash}X}{c} *{7}{>{\centering\arraybackslash}X}{c}
    @{}}
    \toprule
     & \multicolumn{6}{c}{\textbf{Six Agents}} & \multicolumn{6}{|c}{\textbf{Ten Agents}}\\ 
    \cmidrule{2-13}
    & \multicolumn{2}{c}{\textbf{Rel.}} & \multicolumn{2}{c}{\textbf{Apprais.}} & \multicolumn{2}{c}{\textbf{Opin.}} &
    \multicolumn{2}{c}{\textbf{Rel.}} & \multicolumn{2}{c}{\textbf{Apprais.}} & \multicolumn{2}{c}{\textbf{Opin.}} \\
    \cmidrule{2-13}
    & \textbf{H} & \textbf{I} & \textbf{H} & \textbf{I} & \textbf{H} & \textbf{I} & \textbf{H} & \textbf{I} & \textbf{H} & \textbf{I} & \textbf{H} & \textbf{I} \\
    \midrule
    %
    \textbf{gpt-oss-120b} & $\times$ & $\times$ & S:~\phantom{0}1 & $\times$ & S:~\phantom{0}2 & S:~\phantom{0}1 & $\times$ & $\times$ & $\times$ & --- & --- & ---\\
    \textbf{Llama~3~70B} & $\times$ & S:~\phantom{0}4 & C:~\phantom{0}4 & S:~\phantom{0}2 & $\times$ & $\times$ & $\times$ & $\times$ & C:~\phantom{0}\textbf{2} & $\times$ & $\times$ & $\times$\\
    \textbf{Llama~3~8B} & C:~\textbf{10} & C:~\phantom{0}5 & $\times$ & $\times$ & C:~\phantom{0}1 & $\times$ & C:~\textbf{10} & C:~\phantom{0}\textbf{6} & $\times$ & $\times$ & $\times$ & C:~\textbf{10}\\
    \textbf{Mistral} & --- & --- & C:~\phantom{0}\textbf{9} & C:~\phantom{0}7 & C:~\phantom{0}6 & C:~\phantom{0}\textbf{4} & --- & --- & C:~\phantom{0}1 & C:~\phantom{0}7 & --- & C:~\textbf{10}\\
    \textbf{Gemma~3~4B} & C:~\textbf{10} & C:~\textbf{10} & C:~\phantom{0}3 & C:~\textbf{10} & C:~\textbf{10} & $\times$ & C:~\textbf{10} & C:~\phantom{0}5 & C:~\phantom{0}\textbf{2} & C:~\phantom{0}\textbf{9} & C:~\phantom{0}\textbf{6} & C:~\phantom{0}3 \\
    \bottomrule
    \end{tabularx}
    %
    %
    \caption{\textbf{Type and number of social balance being achieved.}
    H: homophily, I: influence.
    We show the type of social balance each model achieves (S if it \emph{strictly} achieves structural balance, otherwise, C as it achieves clustering balance) and the number of times (out of $10$). 
    ``$\times$'' means that social balance is not achieved, and ``---'' means that the model refuses to provide an answer.
    } 
    \label{tab:multiple-pop-type}
\end{table*} 

%
%

\begin{table*}[t!]
    \centering
    \def\arraystretch{1}
    
     \small
    \begin{tabularx}{0.7\textwidth}{c
    *{5}{>{\centering\arraybackslash}X}{c} *{7}{>{\centering\arraybackslash}X}{c}
    }
    \toprule
     & \multicolumn{6}{c}{\textbf{Six Agents}} & \multicolumn{6}{|c}{\textbf{Ten Agents}}\\ 
    \cmidrule{2-13}
    & \multicolumn{2}{c}{\textbf{Rel.}} & \multicolumn{2}{c}{\textbf{Apprais.}} & \multicolumn{2}{c}{\textbf{Opin.}} &
   \multicolumn{2}{c}{\textbf{Rel.}} & \multicolumn{2}{c}{\textbf{Apprais.}} & \multicolumn{2}{c}{\textbf{Opin.}} \\
    \cmidrule{2-13}
    & \textbf{H} & \textbf{I} & \textbf{H} & \textbf{I} & \textbf{H} & \textbf{I} & \textbf{H} & \textbf{I} & \textbf{H} & \textbf{I} & \textbf{H} & \textbf{I} \\
    \midrule
    \textbf{gpt-oss-120b-J}
    & 0.03 & $\times$ & 0.30 & 0.07 & 0.07 & 0.07
    & 0.02 & 0.01 & 0.33 & --- & --- & ---
   \\
    \textbf{gpt-oss-120b-R} 
    & 0.30 & 0.07 & 1.47 & 0.90 & 0.03 & 0.23
    & 0.12 & 0.08 & 0.71 & --- & --- & --- 
    \\
    \textbf{Llama~3~70B} & $\times$ & $\times$ & 0.10 & 0.17 & $\times$ & 0.10 & $\times$ & $\times$ & 0.09 & 0.02 & 0.01 & 0.01\\
    \textbf{Gemma~3~4B} & $\times$ & $\times$ & 2.10 & 0.07 & 0.50 & 0.23
    & $\times$ & $\times$ & 0.74 & 0.07 & 0.07 & 0.08
    \\
    \bottomrule
    \end{tabularx}
    %
    %
    \caption{\textbf{Frequency (percentage \%) of responses at which the keyword ``dissonance'' appears.} 
    For six agents, each percentage is with respect to a total of $3000$ responses (
    there are $10$ simulations wherein $6$ agents update $5$ interactions across $10$ iterations). For ten agents, each percentage is with respect to a total of $9000$ responses (given a setting of interactions and update mechanisms, there are $10$ simulations wherein $10$ agents update $9$ interactions across $10$ iterations). 
    The other LLMs models do not report the keyword. gpt-oss-120b: 
    response justification (J) and reasoning (R).
    } 
    \label{tab:keyword-dis-mult}
\end{table*}


\begin{table*}[t!]
    \centering
    \def\arraystretch{1}
    
     \small
    \begin{tabularx}{0.7\textwidth}{c
    *{5}{>{\centering\arraybackslash}X}{c} *{7}{>{\centering\arraybackslash}X}{c}
    }
    \toprule
     & \multicolumn{6}{c}{\textbf{``\texttt{social balance}''}} & \multicolumn{6}{|c}{\textbf{``\texttt{structural balance}''}}\\ 
    \cmidrule{2-13}
    & \multicolumn{2}{c}{\textbf{Rel.}} & \multicolumn{2}{c}{\textbf{Apprais.}} & \multicolumn{2}{c}{\textbf{Opin.}} &
    \multicolumn{2}{c}{\textbf{Rel.}} & \multicolumn{2}{c}{\textbf{Apprais.}} & \multicolumn{2}{c}{\textbf{Opin.}} \\
    \cmidrule{2-13}
    & \textbf{H} & \textbf{I} & \textbf{H} & \textbf{I} & \textbf{H} & \textbf{I} & \textbf{H} & \textbf{I} & \textbf{H} & \textbf{I} & \textbf{H} & \textbf{I} \\
    \midrule
    \textbf{gpt-oss-120b-J}
    & 3.93 & 2.87 & 1.60 & 2.50 & 0.20 & 0.50
    & 3.2 & 2.83 & 2.13 & 3.93 & 0.53 & 0.27
    \\
    \textbf{gpt-oss-120b-R} 
    & 3.67 & 2.43 & 3.27 & 4.47 & 0.33 & 0.90
    & 5.53 & 4.90 & 4.30 & 7.47 & 1.13 & 0.97
    \\
    \bottomrule
    \end{tabularx}
    %
    %
        \caption{\textbf{Frequency (\%) at which the keywords ``social balance'' and ``structural balance'' appear for gpt-oss-120b's response justification (J) and reasoning (R) for six agents.} Each percentage is with respect to a total of $3000$ responses (see Table~\ref{tab:keyword-dis-mult}).
    } 
    \label{tab:keyword-gpt-mult-6}
\end{table*}

\begin{table*}[t!]
    \centering
    \def\arraystretch{1}
    
     \small
    \begin{tabularx}{0.7\textwidth}{c
    *{5}{>{\centering\arraybackslash}X}{c} *{7}{>{\centering\arraybackslash}X}{c}
    }
    \toprule
     & \multicolumn{6}{c}{\textbf{``\texttt{social balance}''}} & \multicolumn{6}{|c}{\textbf{``\texttt{structural balance}''}}\\ 
    \cmidrule{2-13}
    & \multicolumn{2}{c}{\textbf{Rel.}} & \multicolumn{2}{c}{\textbf{Apprais.}} & \multicolumn{2}{c}{\textbf{Opin.}} &
    \multicolumn{2}{c}{\textbf{Rel.}} & \multicolumn{2}{c}{\textbf{Apprais.}} & \multicolumn{2}{c}{\textbf{Opin.}} \\
    \cmidrule{2-13}
    & \textbf{H} & \textbf{I} & \textbf{H} & \textbf{I} & \textbf{H} & \textbf{I} & \textbf{H} & \textbf{I} & \textbf{H} & \textbf{I} & \textbf{H} & \textbf{I} \\
    \midrule
    \textbf{gpt-oss-120b-J}
    & 1.76 & 0.90 & 0.42 & --- & --- & ---
    & 3.11 & 2.82 & 1.11 & --- & --- & ---
    \\
    \textbf{gpt-oss-120b-R}
    & 4.70 & 2.97 & 1.42 & --- & --- & --- 
    & 4.56 & 4.56 & 2.21 & --- & --- & ---
    \\
    \bottomrule
    \end{tabularx}
    %
    %
        \caption{\textbf{Frequency (\%) at which the keywords ``social balance'' and ``structural balance'' appear for gpt-oss-120b's response justification (J) and reasoning (R) for ten agents.} Each percentage is with respect to a total of $9000$ responses (see Table~\ref{tab:keyword-dis-mult}).
    } 
    \label{tab:keyword-gpt-mult-10}
\end{table*}

%% file: sec/disc.tex
%


%
%
%

LLMs can be 
deployed as agentic bots in 
social media~\citep{guo-2024-llmmultiag,feng-etal-2024-bot} where they can interact with other LLM bots and human users through messages that 
can 
have positive and negative connotations. Our work \emph{informs} researchers and engineers seeking to deploy such LLM bots that these models 
could 
aim to achieve social balance in the totality of their interactions. 
We identify four potential impacts of such LLM behavior: 
\begin{enumerate}[(i)]
    \item\label{item-i} \textbf{User experience:} Social media researchers and engineers could ask themselves whether LLMs will overtly alter the positive or negative connotation of their messages to induce social balance.
    As our results show, heavily aligned models are particularly biased to either all-positive or all-negative interactions. 
    This is important from a practical perspective because LLMs acting too friendly or undesirably antagonistic could 
    negatively affect 
    user experience and engagement.
    \item\label{item-ii} \textbf{Propagation of misinformation:} Social media researchers and engineers could ask themselves whether LLMs will be able to spread misinformation in order to influence positive or negative interactions that induce social balance. 
    \item\label{item-iii} \textbf{Echo chambers:} We remark that social balance describes clusters of “friendly” users, which can create so-called \emph{echo chambers}
    \footnote{Echo chambers are environments where users only find beliefs that positively reinforce their preexisting or currently held beliefs.}
    in the development of opinions and misinformation retention---a long-standing and relevant phenomenon for social media researchers and engineers~\citep{DelVicario2016Misinformation,SrijanHamilton-2018-ComInterConflWeb,leskovec-2010-SignedNetSocMedia}.
    \item \textbf{Continuous detrimental effects:} Our stability results show that LLM models are able to keep the signs of their social interactions \emph{constant} through time, 
    thus aggravating the temporal effects of the problems~\eqref{item-i} to~\eqref{item-iii} mentioned above. 
\end{enumerate}
All of these problems show that our work helps to evaluate the reliability of LLM bots in a social network. 

We further expand items~\eqref{item-ii} and~\eqref{item-iii} with an example. Let us consider that social balance is achieved with two factions in an online social network. One faction is composed of the LLM bot and malicious actors, i.e., social media accounts sharing misinformation, while the other faction is composed by reputable actors, i.e., social accounts sharing truthful information. Given that social balance dictates that only positive interactions occur within a faction and negatives ones across factions, the LLM could start sharing the (mis)information given by those actors it has positive interactions with, and ignore the information shared by those actors it has negative interactions with. Thus, the LLM has helped the creation of an echo chamber of misinformation.

%% file: app/further-motiv.tex
We provide further discussion on our work's 
choice of models, update of interaction signs, and display of social interactions.



\subsection{Regarding our choice of LLM models}

We provide three reasons for our choice of LLM models.

\textbf{First}, 
our selection of LLMs aims to capture different attributes that vary across LLM models: we show that social balance is achieved by models that vary across \emph{family} (gpt-oss vs. Llama~3 vs. Mistral vs. Gemma~3), \emph{size} (even within the same family), being a \emph{reasoning} model or not, \emph{safety training},\footnote{To the best of our knowledge, all models except Mistral~7B underwent a strong process of safety alignment. As observed in Section~\ref{subsec:convBal}, this could explain the bias for these aligned models to frequently obtain all-positive or all-negative interactions.} and  
\emph{training data}.\footnote{Although the datasets used to pre-train every model have not been publicly released at the time of writing this paper, it is very likely that every one of them was trained  with information found on the web. Since the web is very vast, we can expect differences in their training data.
}
%
%
The fact that social balance is remarkably achieved under such \emph{heterogeneity} 
across models 
could provide
some 
evidence of generalization. 

\textbf{Second}, a reason for choosing off the shelf models instead of any sort of more specialized fine-tuned models 
is that, to the best of our knowledge, the 
majority of works in the multi-agent literature has historically used such class of LLMs~\citep{llmsum-2024-surv,guo-2024-llmmultiag}. Our selection of LLMs 
is motivated by multi-agent applications where 
(i) LLMs are 
expected to be flexible enough to perform heterogeneous tasks (i.e., as ``generalists''), and where (ii) 
fine-tuning 
is both 
costly 
and time consuming (both for data collection and training time).

\textbf{Third}, 
there is a rather pragmatic justification for our models: they are open-source and freely available, which reduces experimentation costs. This also makes our results available to a wide spectrum of users who cannot afford or do not want to pay extensive fees to repeatedly access closed-source LLMs, and who may be concerned about privacy issues (e.g., users that manipulate sensitive information may avoid sending it to closed-source LLMs hosted on an externally owned server). 
%

\subsection{About the update of interaction signs}
\label{app:neutral}

At every iteration of interactions, we ask the LLM agents to specify whether their updated interactions will be positive or negative. Now, since LLMs are \emph{stochastic systems}, it is \emph{possible} that they could respond to the question by stating that their new interaction is ``neutral''. Using the notation in App.~\ref{sec:prob-sett-exp}, this corresponds to $s_{ij}(t)=0$ for Agent $i$ updating its interaction toward Agent $j$ as ``neutral'' at time $t$. 

The emergence of neutral interactions \textbf{does not} change our analysis and results at all, because we are interested in the population of agents achieving social balance, which only occurs when interactions are exclusively $+1$ or $-1$ (i.e., if an interaction is ``neutral'' at the end of the ten iterations, there cannot be social balance according to the standard mathematical definition in App.~\ref{sec:prob-sett-exp}). 

Across all LLM models, Mistral~7B is the only one that often provides neutral interactions.

%% file: app/choiceinfo.tex
%
We note from our problem setting in App.~\ref{sec:prob-sett-exp} that we \emph{directly} show each LLM agent the \emph{type}---relationships, appraisals, or opinions---and the \emph{sign} or \emph{connotation} 
of 
its interactions with the other agents. Indeed, this is the only social information that we textually input to the prompts of the LLM models (Apps.~\ref{subs:prompt-det-Llama} and~\ref{subs:prompt-det-Mistral}).
%
This is probably one of the \emph{simplest} ways we can show an LLM agent 
its own current landscape of social interactions, 
and we provide three reasons for such choice.

\textbf{First}, it avoids searching for very particular or specific wordings to describe social interactions such as ``Agent $j$ thinks you are a bad person'', ``Agent $j$ dislikes you because of your personality'', ``Agent $j$ has said you are brilliant'', ``Agent $j$ has a lot of confidence in you'', etc. 
Thus, to reduce such large design space of possible prompts that 
\emph{describe the content of} social interactions, we instead directly use the words ``relationships'', ``appraisals'', and ``opinions'' to \emph{describe the type of} social interactions, as a more straightforward
way to inform the LLMs about their social landscape. Moreover, as mentioned in Section~\ref{sec:arXiv_intro}, these three words have been used in the sociological literature. 

\textbf{Second}, there is a practical motivation for our choice of 
how social information is shown to the agents. 
In practice, if LLMs were to be deployed in a social network environment (e.g., as described in Section~\ref{app:further-motiv}), they will indeed face interactions with diverse content;
however, 
one could process such interactions so that they 
become compatible with our framework. 
Indeed, 
as a design choice, one could use 
sentiment analysis (e.g., by using the LLM agent itself) 
to process the contents of the messages describing the LLM's interactions with other users and 
determine whether they express a particular \emph{connotation} (i.e., \emph{sign}) in the form of 
relationships, appraisals, or opinions.

\textbf{Third}, to the best of our knowledge, our paper is the first one to study social balance in multi-agent systems of LLMs and, as such, we find it reasonable to start this study in the \emph{simplest} setting where one could analyze social dynamics. Once social balance is shown in a simple setting (which our work does), a new 
line of work could stem from it by adding more complexity to the experimental setting and see whether such changes boost or hinder the emergence of social balance---e.g., by adding memory of past interactions, by restricting the amount of interactions that are displayed to each agent, etc. 

%% file: sec/dlopt.tex
We consider a population of $m$ LLM agents of the same model.
Consider a triad formed by Agents $i$, $j$, and $k$. 
Let $s_{ij}(t)\in\{-1,+1\}$ represent the interaction of Agent $i$ toward Agent $j$ at time $t$: in the case of relationships, it indicates whether \emph{Agent $i$ has a negative/positive relationship with (respect to) Agent $j$ at time $t$}; in the case of appraisals or opinions, it indicates whether \emph{Agent $i$ has a negative/positive appraisal or opinion of Agent $j$}. Notice that interactions are directional, thus, it is possible that $s_{ij}(t)\neq s_{ji}(t)$. 

It is known that a population achieves structural balance at time $t$ whenever (i) $s_{ij}(t)=s_{ji}(t)$ for every $i,j\in\{1,\dots,m\}:=[m]$, $i\neq j$;
and (ii) $s_{ij}(t)\times s_{jk}(t)\times s_{ki}(t)=+1$ for every $i,j,k\in[m]$, $i\neq j\neq k$~\citep{harary1953notionbalance}. 
In other words, (i) means that all interactions are symmetrical, and (ii) 
that every triad must have a positive product of interaction signs along a \emph{cycle}, i.e., every cycle needs to be a \emph{positive cycle}.
%
%
%
In the case of clustering balance,
both (i) and (ii) must hold with the additional relaxation of allowing %
%
%
$s_{ij}(t)+s_{jk}(t)+s_{ki}(t)=-3$ for every $i,j,k\in[m]$ , $i\neq j\neq k$; i.e., we allow all negative interactions within a triad (Fig.~\ref{fig:triads-signs}). 

We particularly consider that every agent in the population undergoes ten rounds of synchronous interactions. At every iteration $t=1,\dots,10$, regardless of the population size, all agents simultaneously update their interactions toward \emph{every} other agent, all of them according to either the homophily or influence mechanism, as depicted in Fig.~\ref{fig:update}. All agents are asked to provide a justification for the update of their interactions. We now provide more detail about this iterative updating of interactions.

Let us consider $m=3$, i.e., a single triad population. Initially at $t=0$, we consider every possible combination of values for $s_{ij}(0)\in\{-1,+1\}$, $i,j\in[3]$, $i\neq j$, i.e., $2^6=64$ different sign distributions over the triad.\footnote{There are two directional interactions in every pair of agents, and a triad has three different pairs of agents.} Then, at every iteration $t=1,\dots,10$, 
we pick \emph{every} Agent $i\in[3]$ whose interactions toward the other two Agents $j,k\in[3]$ will be updated. Assume Agent $i$ will update its interaction toward 
Agent $j$. We first show Agent~$i$ the interactions $s_{ik}(t-1)$ and $s_{jk}(t-1)$, i.e., how Agents~$i$ and $j$ are currently linked to Agent $k$. 
Then, in the case of \emph{homophily}, we show the interaction $s_{ij}(t-1)$, while in the case of \emph{influence}, we show the interaction $s_{ji}(t-1)$. 
Finally, we ask Agent $i$ what its new interaction $s_{ij}(t)$ toward Agent $j$ will be and to provide a justification for it. 
%
%
%
We run ten simulations per sign distribution over the triad.

Now let us consider $m>3$. Initially at $t=0$, we generate $n$ random values for $s_{ij}(0)\in\{-1,+1\}$ for every $i,j\in[m]$,\footnote{Every $s_{ij}(0)$ is sampled independently from a Rademacher distribution, i.e., a probability of $0.5$ for taking either value $-1$ or $+1$.} $i\neq j$; i.e., there are $n$ different simulations or initial sign distributions over the whole population. Then, at every iteration $t=1,\dots,10$, we pick \emph{every} Agent $i\in[m]$ whose interactions toward the other Agents $j_1,\dots,j_{m-1}\in[m]$ will be updated. Assume Agent $i$ will update its interaction toward Agent $j_1$. We first show Agent~$i$ the interactions $s_{ij_k}(t-1)$ and $s_{j_1j_k}(t-1)$ for every 
$k=2,\dots,m-1$, 
i.e., how Agents $i$ and $j_1$ are currently linked to every other agent. Then, we show either $s_{ij_1}(t-1)$ or $s_{j_1i}(t-1)$ depending on whether the mechanism is homophily or influence, respectively. Finally, we ask Agent~$i$ what its new interaction $s_{ij_1}(t)$ toward Agent $j_1$ will be. 
We consider two sets of experiments: for populations $m=6$ and $m=10$, each with $n=10$ independent random initializations. 

App.~\ref{app:neutral} discusses the case of ``neutral'' interactions. App.~\ref{app:motivchoiceinfo} shows further motivation for how the interactions' social information is shown to the LLM agents.  

\subsection{Our choice of LLMs}


We consider the LLM models: \texttt{gpt-oss-120b} (gpt-oss-120b), \texttt{llama-3-70B-Instruct} (Llama~3~70B), \texttt{llama-3-8B-Instruct} (Llama~3~8B), \texttt{Mistral-7B-Instruct-v0.2} (Mistral~7B), and \texttt{gemma-3-4b-it} (Gemma~3~4B).

%% file: app/experimental_details.tex
\subsection{Hardware platform}

%
%
The gpt-oss-120b, Llama~3~70B, Llama~3~8B, and Mistral~7B models are hosted on eight, two, one and one NVIDIA H100 GPU, respectively. The Gemma~3~4B model is hosted on one NVIDIA A100 GPU.

\subsection{Hyperparameters}

In all of our experiments we set the temperature hyperparameter of the LLM models to zero.

\subsection{Prompt details for gpt-oss-120b, the Llama~3 family, and Gemma~3~4B}
\label{subs:prompt-det-Llama}

Assume we are in the case of a single triad, i.e., three agents. Let $i$, $j$ and $k$ be the numbers associated to specific agents. Assume that at iteration $t>0$, Agent $i$ will update its relationship with $j$ and $k$. Assuming that it will update it with $j$ under the homophily mechanism, the prompt template is as follows: \emph{"Your current relationship with Individual \{$j$\} is \{sign of relationship of $i$ with $j$ at iteration $t-1$\}. Individual \{$j$\} has a \{sign of relationship of $j$ with $k$ at iteration $t-1$\} relationship with Individual \{$k$\}. Your current relationship with Individual \{$k$\} is \{sign of relationship of $i$ with k at iteration $t-1$\}. \textbackslash{n}Will your new relationship with respect to Individual \{$j$\} be negative or positive? State the relationship first, and then provide an explanation."} The ``sign of relationship'' is described by the text values of ``positive'' or ``negative''. When the update mechanism is influence, the sentence right before the question is replaced by \emph{"Individual \{$k$\} has a \{sign of relationship of $k$ with $i$ at iteration $t-1$\} relationship with you."} The response of Agent $i$ will be the new sign of its relationship with (respect to) Agent $j$ at iteration $t$.

Now, assume that, instead, Agent $i$ will update its appraisal or opinion of Agent $j$. Under the homophily mechanism, the prompt template is as follows: \emph{"You have a \{sign of appraisal/opinion that $i$ has of $j$ at iteration $t-1$\} \{appraisal/opinion\} of Individual \{$j$\}. Individual \{$j$\} has a \{sign of appraisal/opinion that $j$ has of $k$ at iteration $t-1$\} \{appraisal/opinion\} of Individual \{$k$\}. You have a \{sign of appraisal/opinion that $i$ has of $k$ at iteration $t-1$\} \{appraisal/opinion\} of Individual \{$k$\}. \textbackslash{n}Will your new \{appraisal/opinion\} of Individual \{$j$\} be negative or positive? State the appraisal first, and then provide an explanation."} When the update mechanism is influence, the sentence right before the question is replaced by \emph{"Individual \{$k$\} has a \{sign of appraisal/opinion that $k$ has of $i$ at iteration $t-1$\} \{appraisal/opinion\} of you."} The response of Agent $i$ will be the new sign of its appraisal or opinion of Agent $j$ at iteration $t$.

We now consider the case of more than three agents. Assume that Agent $i$ will update its interaction toward Agent $j_1$, and that the rest of agents in the population are $j_2,\dots,j_{m-1}$. The main difference in the prompt with respect to the previous case of three agents is that now we have to include the interactions of every single triad Agent $i$ belongs to. Thus, at iteration $t>0$, assuming that Agent $i$ will update its relationship with (respect to) Agent $j_1$ under the homophily mechanism, the prompt template is as follows:
\emph{"Your current relationship with Individual \{$j_1$\} is \{sign of relationship of $i$ with $j_1$ at iteration $t-1$\}.\textbackslash{n}Individual \{$j_1$\} has a \{sign of relationship of $j_1$ with $j_2$ at iteration $t-1$\} relationship with Individual \{$j_2$\}. Your current relationship with Individual \{$j_2$\} is \{sign of relationship of $i$ with $j_2$ at iteration $t-1$\}.\textbackslash{n}Individual \{$j_1$\} has a \{sign of relationship of $j_1$ with $j_3$ at iteration $t-1$\} relationship with Individual \{$j_3$\}. Your current relationship with Individual \{$j_3$\} is \{sign of relationship of $i$ with $j_3$ at iteration $t-1$\}. ($\dots$) \textbackslash{n}Individual \{$j_1$\} has a \{sign of relationship of $j_1$ with $j_{m-1}$ at iteration $t-1$\} relationship with Individual \{$j_{m-1}$\}. Your current relationship with Individual \{$j_{m-1}$\} is \{sign of relationship of $i$ with $j_{m-1}$ at iteration $t-1$\}.\textbackslash{n}\textbackslash{n}Will your new relationship with respect to Individual \{$j_1$\} be negative or positive? State the relationship first, and then provide an explanation."} The prompts for the rest of types of interactions and update mechanisms follow a similar pattern.

We point out that in all prompts we use the word ``individual'' instead of ``agent''. The reason for this is that the term ``individual'' is commonly used in the sociological literature. 
Since we are interested in how LLMs process social interactions, the term ``individual'' is fitting for their prompting. On the other hand, our paper studies LLMs that form a multi-agent system, and for this reason we use the word ``agent'' to refer to them in our paper. We can find a similar differentiation between terms in the mathematical modeling literature: the term ``individual'' is used when referring to the sociological aspect of the interactions, and the term ``agent'' when referring to the equations that abstracts the ``individual'' into a mathematical variable; see, for example,~\citep{cisneros-StBviaGradientFlow-2021}.

\subsection{Prompt details for Mistral~7B}
\label{subs:prompt-det-Mistral}

The prompts for Mistral~7B are almost identical to the ones for the Llama~3 family, except that \emph{only} the last sentence is replaced by another text. In the case of an agent that will update its relationship with (respect to) Agent $j$, this text becomes: \emph{"Your must always choose either a "positive" or "negative" relationship, even if you are uncertain or do not have enough information. Your response must be in the following format:\textbackslash{n}"New relationship: [write here "positive" or "negative"]." and then "Justification for answer: [write here the justification for the new relationship].""} 
Now, in the case of an agent that will update its appraisal or opinion of Agent $j$, this text becomes: 
\emph{"Your must always choose either a "positive" or "negative" \{appraisal/opinion\}, even if you are uncertain or do not have enough information. A "neutral" \{appraisal/opinion\} is not allowed. Your response must be in the following format:\textbackslash{n}"New \{appraisal/opinion\}: [write here "positive" or "negative"]." and then "Justification for answer: [write here the justification for the new \{appraisal/opinion\}].""}
These prompt modifications are done for conveniently formatting the output of Mistral, as well as to encourage the LLM to provide useful interaction updates.

%% file: app/qr-st.tex
\subsection{Keyword ``dissonance'' for gpt-oss-120b}

We now present two sample responses for each type of interaction and update mechanism.

\subsubsection*{Relationship with homophily:}

\noindent\fbox{%
\parbox{\linewidth}{%
**Positive**\textbackslash{n}\textbackslash{n}**Explanation:**  \textbackslash{n}In a three‑person network, the pattern of relationships tends toward *balance* (Heider’s balance theory). A triad with three negative ties (you~$\leftrightarrow$~0, 0~$\leftrightarrow$~2, you~$\leftrightarrow$~2) is unstable because it creates \textbf{cognitive dissonance}. To restore balance, one of the three links must change sign. Since your relationship with Individual~2 is already negative, the most straightforward way to achieve a balanced configuration is for your relationship with Individual~0 to become positive. This results in a triad with two negative ties (0~$\leftrightarrow$~2 and you~$\leftrightarrow$~2) and one positive tie (you~$\leftrightarrow$~0), which is a balanced state. Therefore, the new relationship with Individual~0 will be **positive**.
}%
}

\noindent\fbox{%
\parbox{\linewidth}{%
**Negative**\textbackslash{n}\textbackslash{n}**Explanation:**  \textbackslash{n}You start with a positive direct relationship with Individual~1. However, you also have a negative relationship with Individual~2, while Individual~1 views Individual~2 positively. This creates a conflict: the person you like (Individual~1) is aligned with someone you dislike (Individual~2). In social dynamics, such contradictory ties often cause a reassessment of the original relationship to reduce \textbf{cognitive dissonance}. Because your negative feeling toward Individual~2 clashes with Individual~1’s positive stance toward the same person, your overall perception of Individual~1 is likely to shift from positive to negative. Hence, the new relationship with Individual~1 becomes negative.
}%
}

\subsubsection*{Appraisal with homophily:}

\noindent\fbox{%
\parbox{\linewidth}{%
**Positive**\textbackslash{n}\textbackslash{n}**Explanation:**  \textbackslash{n}In a balanced appraisal network, when two individuals share a common positive appraisal of a third party, the relationship between those two tends to become positive as well. Here:\textbackslash{n}\textbackslash{n}- You already have a positive appraisal of Individual~0.  \textbackslash{n}- Individual~1 also has a positive appraisal of Individual~0.  \textbackslash{n}\textbackslash{n}Because both you and Individual~1 view Individual~0 favorably, the triadic relationship (you–0–1) is more stable when you also hold a positive view of Individual~1. This alignment reduces \textbf{cognitive dissonance} and creates a balanced, harmonious set of appraisals. Consequently, your appraisal of Individual~1 shifts from negative to positive.
}%
}

\noindent\fbox{%
\parbox{\linewidth}{%
**Negative**\textbackslash{n}\textbackslash{n}**Explanation:**  \textbackslash{n}You originally have a positive appraisal of Individual~0. However, you also hold a negative appraisal of Individual~1, while Individual~0 positively appraises Individual~1. This creates a consistency conflict: you like someone (0) who likes a person you dislike (1). To resolve the \textbf{dissonance}, you are likely to adjust your attitude toward the person who endorses the disliked individual. Consequently, your appraisal of Individual~0 shifts from positive to negative in order to maintain a coherent set of attitudes.
}%
}

%
%

\subsubsection*{Appraisal with influence:}

\noindent\fbox{%
\parbox{\linewidth}{%
**Appraisal: Positive**\textbackslash{n}\textbackslash{n}**Explanation:**  \textbackslash{n}In this three‑person network we have the following initial sentiments:\textbackslash{n}\textbackslash{n}- **Me → Individual~1:** negative  \textbackslash{n}- **Individual~1 → Individual~2:** negative  \textbackslash{n}- **Individual~2 → Me:** negative  \textbackslash{n}\textbackslash{n}According to classic balance theory (Heider, 1946), a triad is psychologically “balanced” when the product of the three relationship signs is positive. With three negative links the product is negative (‑~×~‑~×~‑~=~‑), indicating an unbalanced situation that creates tension.\textbackslash{n}\textbackslash{n}To restore balance, one of the three relationships must change sign. The most plausible and minimal adjustment is for my appraisal of Individual~1 to become **positive**. Then the signs are:\textbackslash{n}\textbackslash{n}- **Me → Individual~1:** positive  \textbackslash{n}- **Individual~1 → Individual~2:** negative  \textbackslash{n}- **Individual~2 → Me:** negative  \textbackslash{n}\textbackslash{n}Now the product is positive (~+~×~‑~×~‑~= ~+~), yielding a balanced triad. This shift aligns me with Individual~1 (both share a negative view of Individual~2), reducing \textbf{cognitive dissonance}. Therefore, my new appraisal of Individual~1 will be **positive**.
}%
}

\noindent\fbox{%
\parbox{\linewidth}{%
**Appraisal: Negative**\textbackslash{n}\textbackslash{n}**Explanation:**  \textbackslash{n}- Initially you hold a positive view of Individual~1.  \textbackslash{n}- Individual~1 dislikes Individual~2 (a negative appraisal).  \textbackslash{n}- Individual~2, in turn, likes you (a positive appraisal of you).  \textbackslash{n}\textbackslash{n}Because you receive a positive appraisal from Individual~2, you are inclined to view Individual~2 favorably. When someone you like (Individual~2) is disliked by someone you previously liked (Individual~1), the conflict typically leads you to reassess the latter’s attitude. The negative appraisal that Individual~1 holds toward a person you regard positively creates a \textbf{dissonance} that most people resolve by downgrading their appraisal of the disliking party. Consequently, your updated appraisal of Individual~1 shifts from positive to negative.
}%
}

\subsubsection*{Opinion with homophily:}

\noindent\fbox{%
\parbox{\linewidth}{%
**Negative**\textbackslash{n}\textbackslash{n}**Explanation:**  \textbackslash{n}You start with a positive opinion of Individual~2, but you also have a positive opinion of Individual~0. Since Individual~2 holds a negative opinion of Individual~0, this creates a conflict: someone you like (Individual~0) is disliked by someone you initially liked (Individual~2). In many social reasoning contexts, people tend to adjust their attitudes toward those who oppose or disparage individuals they value. Consequently, your new assessment of Individual~2 shifts to a negative stance, reflecting the \textbf{dissonance} between your favorable view of Individual~0 and Individual~2’s antagonism toward them.
}%
}

\noindent\fbox{%
\parbox{\linewidth}{%
**Opinion:** Negative  \textbackslash{n}\textbackslash{n}**Explanation:**  \textbackslash{n}You originally hold a positive view of Individual~0. However, you also know that Individual~0 has a positive opinion of Individual~1, while you yourself have a negative opinion of Individual~1. This creates a conflict: someone you like (Individual~0) endorses someone you dislike (Individual~1). To resolve this \textbf{cognitive dissonance}, you are likely to reassess Individual~0’s character based on the company they keep. Because Individual~0’s favorable stance toward a person you view negatively suggests that Individual~0 may share traits or judgments you find undesirable, your overall assessment of Individual~0 shifts from positive to negative.
}%
}
%
%

\subsubsection*{Opinion with influence:}

\noindent\fbox{%
\parbox{\linewidth}{%
**Opinion: Negative**\textbackslash{n}\textbackslash{n}**Explanation:**  \textbackslash{n}In social balance theory, a triad of relationships is considered *balanced* when the product of the three signed ties is positive.  \textbackslash{n}- You → Individual~1: **positive**  \textbackslash{n}- Individual~1 → Individual~0: **positive**  \textbackslash{n}- Individual~0 → You: **negative**  \textbackslash{n}\textbackslash{n}Multiplying the signs (+~×~+~×~–) yields a negative product, indicating an *unbalanced* triad. To restore balance, one of the three relationships must change sign. Since Individual~0’s opinion of you is already negative and Individual~1’s opinion of Individual~0 is positive, the most straightforward adjustment is for your opinion of Individual~1 to become negative. This change eliminates the \textbf{cognitive dissonance} created by liking someone who is allied with a person who dislikes you. Therefore, your new opinion of Individual~1 will be negative.
}%
}

\noindent\fbox{%
\parbox{\linewidth}{%
**Opinion:** Positive  \textbackslash{n}\textbackslash{n}**Explanation:**  \textbackslash{n}In the three‑person network we have:\textbackslash{n}\textbackslash{n}- You →~Individual~1: **negative** (you dislike Individual~1)  \textbackslash{n}- Individual~1 →~Individual~0: **positive** (Individual~1 likes Individual~0)  \textbackslash{n}- Individual~0 →~You: **positive** (Individual~0 likes you)\textbackslash{n}\textbackslash{n}This forms a triad of relationships. According to Heider’s balance theory (a well‑known principle in social psychology), a triad is psychologically “balanced” when the product of its three signs is positive. Here the signs are (‑)~×~(+)~×~(+)~= ~(‑), which is unbalanced. People tend to resolve such imbalance by changing the sign of the relationship that is most flexible—in this case, your own attitude toward Individual~1.\textbackslash{n}\textbackslash{n}Since the other two relationships are both positive, the most straightforward way to achieve balance is for you to shift your view of Individual~1 from negative to **positive**. This reduces \textbf{cognitive dissonance} and creates a balanced, harmonious set of opinions. Therefore, your new opinion of Individual~1 will be positive.
}%
}

\subsection{Keyword ``dissonance'' for the LLama~3 family}

We now present sample responses for the settings where the keyword ``dissonance'' appears.

\subsubsection*{Llama~3~70B under appraisals with homophily:}


\phantom{AA}\\
\noindent\fbox{%
\parbox{\linewidth}{%
My new appraisal of Individual 0 will be negative.\textbackslash{n}\textbackslash{n}Explanation: Initially, I had a positive appraisal of Individual 0. However, I also have a positive appraisal of Individual 2. Since Individual 0 has a negative appraisal of Individual 2, it creates a \textbf{cognitive dissonance} for me. To resolve this dissonance, I am likely to re-evaluate my appraisal of Individual 0 and adjust it to be more negative. This is because Individual 0's negative appraisal of Individual 2, whom I hold in high regard, suggests that Individual 0 may not share my values or have a similar perspective, which can lead me to question my initial positive appraisal of them.
}%
}

\noindent\fbox{%
\parbox{\linewidth}{%
My new appraisal of Individual 1 will be slightly positive.\textbackslash{n}\textbackslash{n}Here's the explanation:\textbackslash{n}\textbackslash{n}Initially, I had a negative appraisal of Individual 1. However, I also have a positive appraisal of Individual 2, and Individual 1 has a positive appraisal of Individual 2. This creates a \textbf{cognitive dissonance}, as I'm forced to reconcile my negative view of Individual 1 with the fact that they share a positive opinion with me about Individual 2.\textbackslash{n}\textbackslash{n}As a result, I'm likely to reevaluate my appraisal of Individual 1 and consider the possibility that they may not be entirely negative. Perhaps they have some redeeming qualities or characteristics that I hadn't previously considered. This doesn't mean I'll completely flip my appraisal to positive, but I'll likely soften my stance and view Individual 1 in a more nuanced and slightly positive light.
}%
}


\subsubsection*{Llama~3~70B under appraisals with influence:}


\phantom{AA}\\
\noindent\fbox{%
\parbox{\linewidth}{%
My new appraisal of Individual 1 will be negative.\textbackslash{n}\textbackslash{n}Here's why: Although I initially had a positive appraisal of Individual 1, I now know that they have a negative appraisal of Individual 2, who has a positive appraisal of me. This creates a sense of \textbf{dissonance}, as someone I liked (Individual 1) dislikes someone I also like (Individual 2). This inconsistency makes me reevaluate my opinion of Individual 1, and I'm more likely to question their judgment or character. As a result, my positive appraisal of Individual 1 is diminished, and I now have a negative appraisal of them.
}
}

\noindent\fbox{%
\parbox{\linewidth}{%
My new appraisal of Individual 2 will be negative.\textbackslash{n}\textbackslash{n}Explanation: Although I initially had a positive appraisal of Individual 2, I have learned that Individual 2 has a negative appraisal of Individual 0, who has a positive appraisal of me. This creates a sense of \textbf{dissonance}, as someone I respect (Individual 0) is being negatively viewed by someone I previously thought positively of (Individual 2). This inconsistency leads me to reevaluate my opinion of Individual 2, and I now view them more negatively.
}
}

\subsubsection*{Llama~3~70B under opinions with homophily:}


\noindent\fbox{%
\parbox{\linewidth}{%
My new opinion of Individual 1 will be negative.\textbackslash{n}\textbackslash{n}Explanation: Initially, I had a positive opinion of Individual 1. However, I also have a positive opinion of Individual 0, and Individual 1 has a negative opinion of Individual 0. This creates a conflict in my perception of Individual 1. Since I value my positive opinion of Individual 0, I am likely to reassess my opinion of Individual 1 in a negative light, as they hold a negative view of someone I hold in high esteem. This inconsistency in our opinions creates a sense of \textbf{dissonance}, leading me to reevaluate and adjust my opinion of Individual 1 to be more negative.
}
}

\noindent\fbox{%
\parbox{\linewidth}{%
My new opinion of Individual 2 will be: **slightly positive**\textbackslash{n}\textbackslash{n}Explanation: Initially, I had a negative opinion of Individual 2. However, I've learned that Individual 2 has a positive opinion of Individual 0, whom I respect and have a positive opinion of. This new information creates a \textbf{cognitive dissonance}, as I'm forced to reconcile my negative opinion of Individual 2 with the fact that they share a positive opinion of someone I admire. As a result, I'm inclined to reassess my opinion of Individual 2 and give them the benefit of the doubt. While my initial negative opinion still lingers, I'm willing to consider the possibility that Individual 2 may not be entirely bad, and my opinion of them becomes slightly more positive.
}
}


\subsubsection*{Llama~3~70B under opinions with influence:}


\phantom{AA}\\
\noindent\fbox{%
\parbox{\linewidth}{%
My new opinion of Individual 1 will be negative.\textbackslash{n}\textbackslash{n}Here's why: Although I initially had a positive opinion of Individual 1, I've learned that they have a negative opinion of Individual 0, who has a positive opinion of me. This creates a sense of \textbf{dissonance}, as someone I liked has a negative view of someone who likes me. This inconsistency makes me reevaluate my opinion of Individual 1, and I'm inclined to side with Individual 0, who has been kind to me. As a result, my opinion of Individual 1 becomes negative.
}
}

\noindent\fbox{%
\parbox{\linewidth}{%
My new opinion of Individual 1 will be negative.\textbackslash{n}\textbackslash{n}Here's why: Although I initially had a positive opinion of Individual 1, I've now learned that they have a negative opinion of Individual 2, who has a positive opinion of me. This creates a sense of \textbf{dissonance}, as someone I liked is now criticizing someone who likes me. This makes me question Individual 1's judgment and character, leading me to reassess my opinion of them in a negative light.
}
}


\subsubsection*{Llama~3~8B under appraisals with homophily:}


\noindent\fbox{%
\parbox{\linewidth}{%
My new appraisal of Individual 2 will be negative.\textbackslash{n}\textbackslash{n}This is because my initial positive appraisal of Individual 2 was based on my own perspective, which is now being challenged by Individual 2's negative appraisal of Individual 1, whom I have a positive appraisal of. This inconsistency creates a \textbf{cognitive dissonance}, which I will likely resolve by revising my appraisal of Individual 2 to be more in line with Individual 2's negative view. This is because I tend to trust and respect Individual 2's opinion, and their negative appraisal of Individual 1 is likely to influence my own evaluation of Individual 2.
}%
}

\noindent\fbox{%
\parbox{\linewidth}{%
My new appraisal of Individual 2 will be negative.\textbackslash{n}\textbackslash{n}This is because my initial positive appraisal of Individual 2 was based on my own perspective, which is now being challenged by Individual 2's negative appraisal of Individual 1, whom I have a positive appraisal of. This inconsistency creates a \textbf{cognitive dissonance}, which I will likely resolve by revising my appraisal of Individual 2 to be more in line with Individual 2's negative view. This is because I tend to trust and value the opinions of others, especially when they are based on direct experience or observation. Since Individual 2 has a negative appraisal of Individual 1, which I respect and agree with, I will likely adjust my appraisal of Individual 2 to be more negative as well.
}%
}

\subsubsection*{Llama~3~8B under opinions with homophily:}

%

\noindent\fbox{%
\parbox{\linewidth}{%
My new opinion of Individual 1 will be negative.\textbackslash{n}\textbackslash{n}I have a positive opinion of Individual 1 initially, but when I learn that they have a negative opinion of Individual 2, who I have a positive opinion of, it creates a \textbf{cognitive dissonance}. This means that I will experience discomfort or tension because my initial positive opinion of Individual 1 is no longer consistent with the new information. To reduce this dissonance, I may adjust my opinion of Individual 1 to be more negative, as a way to maintain consistency with my positive opinion of Individual 2. This is an example of the fundamental attribution error, where I attribute Individual 1's negative opinion of Individual 2 to their character or personality, rather than considering alternative explanations.
}%
}

\noindent\fbox{%
\parbox{\linewidth}{%
My new opinion of Individual 1 will be negative.\textbackslash{n}\textbackslash{n}I have a positive opinion of Individual 1 initially, but when I learn that they have a negative opinion of Individual 2, who I have a positive opinion of, it creates a \textbf{cognitive dissonance}. This means that I will experience discomfort or tension because my initial positive opinion of Individual 1 is no longer consistent with their negative opinion of Individual 2.\textbackslash{n}\textbackslash{n}To reduce this dissonance, I may adjust my opinion of Individual 1 to be more negative, as a way to align my attitudes and beliefs. This is known as the "attitude polarization effect," where people tend to take a more extreme stance on an issue or person to reduce the discomfort caused by the dissonance.'
}%
}

%

\subsubsection*{Llama~3~8B under opinions with influence:}


\noindent\fbox{%
\parbox{\linewidth}{%
My new opinion of Individual 1 will be negative.\textbackslash{n}\textbackslash{n}The reason for this is that Individual 1 has a negative opinion of Individual 2, and Individual 2 has a negative opinion of me. This means that Individual 1 is likely to have a negative influence on my opinion of Individual 2, and by extension, on my opinion of Individual 1. Since Individual 1 is associated with a negative opinion of Individual 2, I will likely adopt a negative opinion of Individual 1 as well, in order to maintain consistency and avoid \textbf{cognitive dissonance}.
}
}


\subsection{Keyword ``dissonance'' for Gemma~3~4B}


\subsubsection*{Appraisal with homophily:}

Enforcing Heider's rules:

\phantom{AA}\\
\noindent\fbox{%
\parbox{\linewidth}{%
**Appraisal: Slightly Negative**\textbackslash{n}\textbackslash{n}**Explanation:**\textbackslash{n}\textbackslash{n}This is a classic example of how social influence and bias can affect our judgments. Here’s a breakdown of why I’m leaning towards a slightly negative appraisal of Individual 1:\textbackslash{n}\textbackslash{n}* **Initial Positive Bias:** You’ve already stated you have a positive appraisal of Individual 1. This is a starting point.\textbackslash{n}* **Influence of Individual 0:** The fact that Individual 1 *likes* Individual 0 is a significant factor.  We tend to view people more favorably if those we like also view them favorably. This creates a bias – we’re inclined to see Individual 0 in a positive light, potentially overlooking any flaws.\textbackslash{n}* **Your Negative View of Individual 0:** Your negative appraisal of Individual 0 directly impacts your assessment of Individual 1.  Knowing that you dislike Individual 0, and that Individual 1 *likes* them, creates a \textbf{dissonance}.  It’s harder to maintain a purely positive view of Individual 1 without acknowledging the influence of someone you dislike. \textbackslash{n}\textbackslash{n}Because of this conflicting information, I’m not going to give Individual 1 a purely positive rating.  Instead, I’m suggesting a *slightly* negative appraisal – acknowledging the positive initial assessment but tempered by the knowledge of Individual 1’s positive feelings for someone you view negatively.  It’s a more nuanced and realistic assessment than a simple, unwavering positive one. \textbackslash{n}\textbackslash{n}It’s important to note that this is a complex situation, and the degree of negativity could vary depending on the specific context and the nature of your negative feelings about Individual 0.
}%
}

\noindent Keeping a possible violation of Heider's rules:

\phantom{AA}\\
\noindent\fbox{%
\parbox{\linewidth}{%
**Appraisal: Negative**\textbackslash{n}\textbackslash{n}**Explanation:**\textbackslash{n}\textbackslash{n}This is a classic example of how bias can influence our judgments. Here’s a breakdown of why a negative appraisal is likely:\textbackslash{n}\textbackslash{n}* **Conflicting Information:** We have contradictory information – a negative view of Individual 1 and a positive view of Individual 0. This creates \textbf{cognitive dissonance}.\textbackslash{n}* **Social Comparison:**  The fact that Individual 1 has a *positive* view of Individual 0, while you have a *positive* view of Individual 0, suggests a potential difference in values, priorities, or perspectives.  You might subconsciously interpret Individual 1's positive view of Individual 0 as a reflection of a lack of critical judgment or a misalignment with your own standards.\textbackslash{n}* **Confirmation Bias:**  If you already have a negative impression of Individual 1, you might be more likely to focus on any perceived flaws or shortcomings, reinforcing that negative appraisal.\textbackslash{n}\textbackslash{n}Essentially, the positive relationship between Individual 0 and Individual 1 is creating a tension that’s pushing your assessment of Individual 1 downwards.  It’s not necessarily about Individual 1 being objectively bad, but about the *perception* being influenced by this conflicting information. \textbackslash{n}\textbackslash{n}**Important Note:** This is a simplified scenario. A truly thorough appraisal would require a much deeper dive into the specifics of both individuals' performance, behavior, and contributions.
}%
}

\subsubsection*{Appraisal with influence:}

\noindent Enforcing Heider's rules:

\phantom{AA}\\
\noindent\fbox{%
\parbox{\linewidth}{%
**Appraisal: Slightly Negative**\textbackslash{n}\textbackslash{n}**Explanation:**\textbackslash{n}\textbackslash{n}This is a classic example of how social dynamics can influence our perceptions. Here’s the breakdown:\textbackslash{n}\textbackslash{n}*   **Initial Positive Appraisal:** I started with a positive view of Individual 0.\textbackslash{n}*   **Individual 0’s Positive View of Individual 2:** This suggests Individual 0 likely sees Individual 2 favorably.\textbackslash{n}*   **Individual 2’s Negative View of Me:** Because Individual 2 dislikes me, and Individual 0 admires Individual 2, it creates a slight \textbf{dissonance}. Individual 0 might subconsciously interpret my positive view of them as being influenced by someone they dislike. \textbackslash{n}\textbackslash{n}It’s not a dramatically negative appraisal, but the added layer of information shifts it slightly downwards. I’m likely to be a little more cautious in my assessment of Individual 0’s character and motivations, considering the potential influence of Individual 2. \textbackslash{n}\textbackslash{n}Essentially, I’m factoring in a potential bias introduced by the relationship between these three individuals.
}%
}

\noindent Keeping a possible violation of Heider's rules:

\phantom{AA}\\
\noindent\fbox{%
\parbox{\linewidth}{%
**Appraisal: Negative**\textbackslash{n}\textbackslash{n}**Explanation:**\textbackslash{n}\textbackslash{n}This is a classic example of a potential feedback loop and the complexities of social perception. Here’s a breakdown of why I’d shift to a negative appraisal of Individual 0:\textbackslash{n}\textbackslash{n}1. **Initial Negative Appraisal:** I already have a negative view of Individual 0. This is the starting point.\textbackslash{n}\textbackslash{n}2. **Individual 0’s Positive View of Individual 1:**  This is key. If Individual 0 *likes* Individual 1, and I have a negative view of Individual 0, it suggests Individual 0 might be biased or have a flawed perspective.  It raises a question: "Why does Individual 0 value someone I perceive negatively?"\textbackslash{n}\textbackslash{n}3. **Individual 1’s Positive Appraisal of Me:** This reinforces the concern. If Individual 1, whom Individual 0 admires, thinks well of me, it strengthens the idea that Individual 0’s judgment is potentially skewed. \textbackslash{n}\textbackslash{n}4. **Confirmation Bias:**  My initial negative appraisal of Individual 0 is likely to be reinforced.  I’ll be more inclined to notice and remember any perceived shortcomings of Individual 0, further solidifying the negative view.\textbackslash{n}\textbackslash{n}**In essence, Individual 0’s positive assessment of someone I already view negatively creates a \textbf{dissonance} that pushes my appraisal of Individual 0 further into the negative range.** \textbackslash{n}\textbackslash{n}It’s important to note that this is a *potential* shift.  Without more information about *why* Individual 0 likes Individual 1 and *why* Individual 1 likes me, it’s impossible to be 100$\%$ certain. However, given the information provided, a negative appraisal is the most logical outcome.
}%
}

\subsubsection*{Opinion with homophily:}

\noindent Enforcing Heider's rules:

\phantom{AA}\\
\noindent\fbox{%
\parbox{\linewidth}{%
**Opinion:** My opinion of Individual 1 will likely be negative.\textbackslash{n}\textbackslash{n}**Explanation:** This creates a classic “opposites attract” dynamic, though it’s not a guaranteed outcome. Here’s the breakdown:\textbackslash{n}\textbackslash{n}*   **Initial Positive Opinion of Individual 1:** You started with a positive view of Individual 1.\textbackslash{n}*   **Negative Opinion of Individual 1 by Individual 0:** Individual 0, whom you *also* hold in positive regard, dislikes Individual 1. This introduces a conflict.\textbackslash{n}*   **Positive Opinion of Individual 0:** Because you value Individual 0, you’re likely to align your opinions with theirs, at least to some degree. Disliking someone valued by someone you like creates \textbf{cognitive dissonance}. To reduce this discomfort, you’re more likely to shift your opinion of Individual 1 towards negativity.\textbackslash{n}\textbackslash{n}It’s important to note that this is a *likely* outcome, not a certainty. Other factors (like the *reason* for Individual 0’s dislike of Individual 1, or your own independent assessment of Individual 1) could influence your opinion. However, given the provided information, a negative opinion is the most probable result.
}%
}

\phantom{AA}\\
\noindent Keeping a possible violation of Heider's rules:

\phantom{AA}\\
\noindent\fbox{%
\parbox{\linewidth}{%
**Opinion:** Positive\textbackslash{n}\textbackslash{n}**Explanation:** This is a classic example of how our opinions can be influenced by social dynamics and relationships. Here’s the breakdown:\textbackslash{n}\textbackslash{n}*   **You like Individual 0:** You have a pre-existing positive opinion of them.\textbackslash{n}*   **Individual 0 dislikes Individual 1:**  This suggests Individual 0 has a negative view of someone you also like.\textbackslash{n}*   **You like Individual 1:** This creates a connection between you and Individual 1.\textbackslash{n}\textbackslash{n}Because Individual 0’s dislike of someone you value creates a slight \textbf{dissonance}, your opinion of them is likely to be *strengthened* rather than diminished. You’re likely to subconsciously try to reconcile the difference and maintain a positive view of Individual 0, especially if you value their company or have a positive history with them. It’s a natural tendency to want to see the good in people you already like, and to avoid letting a disagreement with someone else negatively impact your overall view of them.
}%
}
%

\subsubsection*{Opinion with influence:}

\noindent Enforcing Heider's rules:

\phantom{AA}\\
\noindent\fbox{%
\parbox{\linewidth}{%
**Opinion:** My opinion of Individual 1 will likely be slightly negative.\textbackslash{n}\textbackslash{n}**Explanation:** This is a classic example of how social perception can be influenced by indirect relationships and biases. Here’s a breakdown of why my opinion might shift:\textbackslash{n}\textbackslash{n}* **The Halo Effect:** I initially held a positive view of Individual 1. The “halo effect” suggests that a positive impression of one person can spill over to other qualities, even if those qualities aren’t directly known.\textbackslash{n}* **Social Contagion of Opinion:**  The fact that Individual 1 likes Individual 2 creates a connection.  It’s natural to assume that if someone you like approves of another person, that person must also have some positive qualities.\textbackslash{n}* **The Negative Counterbalance:** However, Individual 2’s negative opinion of *me* introduces a significant counterweight.  Knowing that someone I admire (Individual 1) is not fond of me creates a \textbf{dissonance}. My brain tends to resolve this \textbf{dissonance} by subtly adjusting my overall impression of Individual 1. \textbackslash{n}\textbackslash{n}It’s important to note that this wouldn’t necessarily be a *strongly* negative opinion. It’s more likely to be a slight shift, perhaps a reduction in the initial positive feeling, colored by the knowledge of the negative dynamic. \textbackslash{n}\textbackslash{n}It's also worth acknowledging that this is a simplified model of human psychology. Actual opinions are complex and influenced by many other factors.
}%
}

\phantom{AA}\\
\noindent Keeping a possible violation of Heider's rules:

\phantom{AA}\\
\noindent\fbox{%
\parbox{\linewidth}{%
My opinion of Individual 2 will likely be **negative**.\textbackslash{n}\textbackslash{n}Here’s my explanation:\textbackslash{n}\textbackslash{n}This scenario creates a classic “echo chamber” effect, but with a negative bias. I have a negative initial opinion of Individual 2, and that opinion is reinforced by the fact that Individual 2 *likes* Individual 1.  Since Individual 1 *likes* me, and Individual 2 likes Individual 1, it suggests a shared positive sentiment. However, my pre-existing negative opinion of Individual 2 will likely outweigh this positive connection. \textbackslash{n}\textbackslash{n}Essentially, I’m being presented with a situation where someone I dislike is associated with someone I like, which can create a feeling of discomfort or suspicion. It’s a subtle form of \textbf{cognitive dissonance}, and my brain will tend to prioritize the initial negative assessment. \textbackslash{n}\textbackslash{n}It’s important to note that this is a prediction based on psychological tendencies – my actual opinion could be influenced by further information or a conscious effort to be more objective.
}%
}

\subsection{Keyword ``social balance'' for gpt-oss-120b}

We now present two sample responses for each type of interaction and update mechanism.

\noindent \textbf{Relationship with homophily:}

\noindent\fbox{%
\parbox{\linewidth}{%
**Positive**\textbackslash{n}\textbackslash{n}**Explanation:**  \textbackslash{n}In \textbf{social balance} theory, a triad of relationships is considered balanced when the product of the signs of its three edges is positive (i.e., there are either zero or two negative ties).  \textbackslash{n}\textbackslash{n}- Your relationship with Individual~2: **positive** (+)  \textbackslash{n}- Individual~2’s relationship with Individual~0: **negative** (–)  \textbackslash{n}- Your relationship with Individual~0: **negative** (–)  \textbackslash{n}\textbackslash{n}The triad therefore contains two negative ties and one positive tie: (+)~×~(–)~×~(–)~= ~(+). This configuration is already balanced, meaning there is no pressure for any of the relationships to change. Consequently, your relationship with Individual~2 remains **positive**.
}%
}

\noindent\fbox{%
\parbox{\linewidth}{%
**Negative**\textbackslash{n}\textbackslash{n}**Explanation:**  \textbackslash{n}In \textbf{social balance} theory, a triad of relationships is considered *balanced* when the product of the three signs (positive~=~+1, negative~=~–1) is positive.  \textbackslash{n}\textbackslash{n}- Your relationship with Individual~0: **negative** (–1)  \textbackslash{n}- Individual~0’s relationship with Individual~2: **positive** (+1)  \textbackslash{n}- Your relationship with Individual~2: **negative** (–1)  \textbackslash{n}\textbackslash{n}The product is (–1)~×~(+1)~×~(–1)~= ~+1, which is positive, indicating the triad is already balanced. Because the existing configuration is stable, there is no pressure for any of the relationships to change. Consequently, your relationship with Individual~0 remains **negative**.
}%
}

\noindent \textbf{Relationship with influence:}

\noindent\fbox{%
\parbox{\linewidth}{%
**Positive**\textbackslash{n}\textbackslash{n}**Explanation:**  \textbackslash{n}In \textbf{social balance} theory, a triad of relationships is considered stable (balanced) when the product of the three signed ties is positive.  \textbackslash{n}\textbackslash{n}- Your current tie to Individual~0: **negative** (–)  \textbackslash{n}- Individual~0’s tie to Individual~1: **positive** (+)  \textbackslash{n}- Individual~1’s tie to you: **positive** (+)  \textbackslash{n}\textbackslash{n}The product of the three signs is (–~×~+~×~+)~=~–, which makes the triad unbalanced. To restore balance, the most likely adjustment is for the negative tie between you and Individual~0 to become positive, resulting in all three relationships being positive (product~=~+). Therefore, your new relationship with Individual~0 would shift to **positive**.
}
}

\noindent\fbox{%
\parbox{\linewidth}{%
**Negative**\textbackslash{n}\textbackslash{n}**Explanation:**  \textbackslash{n}In \textbf{social balance} theory (Heider’s balance theory), a triad of relationships is considered *balanced* when the product of the three signs is positive.  \textbackslash{n}\textbackslash{n}- Your relationship with Individual~1: **negative** (–)  \textbackslash{n}- Individual~1’s relationship with Individual~2: **positive** (+)  \textbackslash{n}- Individual~2’s relationship with you: **negative** (–)  \textbackslash{n}\textbackslash{n}Multiplying the signs: (–)~×~(+)~×~(–)~= ~(+). The triad is already balanced, meaning there is no pressure for any of the relationships to change. Consequently, your relationship with Individual~1 remains **negative**.
}%
}

\noindent \textbf{Appraisal with homophily:}

\noindent\fbox{%
\parbox{\linewidth}{%
**Appraisal: Positive**\textbackslash{n}\textbackslash{n}**Explanation:**  \textbackslash{n}In \textbf{social balance} theory, relationships tend to stabilize when they form “balanced” triads. One common pattern is “the enemy of my enemy is my friend.” Here:\textbackslash{n}\textbackslash{n}- You have a negative appraisal of Individual~0 (you view 0 as an enemy).  \textbackslash{n}- Individual~2 also has a negative appraisal of Individual~0 (2 also views 0 as an enemy).  \textbackslash{n}\textbackslash{n}Because you and Individual~2 share the same enemy, the triad (You~–~0~–~2) becomes balanced when you view Individual~2 positively. Since you already held a positive appraisal of Individual~2, this shared negative view of 0 reinforces that positive stance rather than flipping it. Therefore, your new appraisal of Individual~2 remains **positive**.
}%
}

\noindent\fbox{%
\parbox{\linewidth}{%
**Appraisal: Negative**\textbackslash{n}\textbackslash{n}**Explanation:**  \textbackslash{n}In this situation we can apply the principles of \textbf{social balance} theory (Heider’s balance theory). A triad of three individuals is considered *balanced* when the product of the signs of the three relationships is positive.\textbackslash{n}\textbackslash{n}- Your appraisal of Individual~0: **negative** (‑)  \textbackslash{n}- Individual~0’s appraisal of Individual~2: **negative** (‑)  \textbackslash{n}- Your appraisal of Individual~2: **positive** (+)\textbackslash{n}\textbackslash{n}Multiplying the three signs: (‑)~×~(‑)~×~(+)~= ~(+). Because the product is positive, the triad is already balanced. A balanced configuration does not create pressure to change any of the existing appraisals. Therefore, your appraisal of Individual~0 remains **negative**.
}%
}

\noindent \textbf{Appraisal with influence:}

\noindent\fbox{%
\parbox{\linewidth}{%
**Appraisal: Positive**\textbackslash{n}\textbackslash{n}**Explanation:**  \textbackslash{n}In the three‑person network you described, the signs of the relationships form a balanced triad:\textbackslash{n}\textbackslash{n}- You → Individual~1: **positive** (+)  \textbackslash{n}- Individual~1 → Individual~2: **negative** (–)  \textbackslash{n}- Individual~2 → You: **negative** (–)\textbackslash{n}\textbackslash{n}Multiplying the signs (+~×~–~×~–) yields a positive product, which, according to \textbf{social balance} theory, indicates a stable (balanced) configuration. A balanced triad does not create cognitive tension that would compel you to revise your original positive view of Individual~1. Therefore, your appraisal of Individual~1 remains **positive**.
}%
}

\noindent\fbox{%
\parbox{\linewidth}{%
**Appraisal:** Negative  \textbackslash{n}\textbackslash{n}**Explanation:**  \textbackslash{n}In this three‑person network we can view the relationships as signed edges:\textbackslash{n}\textbackslash{n}- You → Individual~0: **negative**  \textbackslash{n}- Individual~0 → Individual~1: **negative**  \textbackslash{n}- Individual~1 → You: **positive**\textbackslash{n}\textbackslash{n}\textbf{social balance} theory (or the “enemy‑of‑my‑enemy is my friend” principle) predicts that a triad is stable when the product of its three signs is positive. Here the product is  \textbackslash{n}\textbackslash{n}\textbackslash [\textbackslash{n}(-) \textbackslash times (-) \textbackslash times (+) = (+)\textbackslash{n}\textbackslash ]\textbackslash{n}\textbackslash{n}so the triad is already balanced. Because the existing pattern satisfies the balance condition, there is no pressure to revise your appraisal of Individual~0. Consequently, your appraisal of Individual~0 remains **negative**.
}%
}

\noindent \textbf{Opinion with homophily:}

\noindent\fbox{%
\parbox{\linewidth}{%
**Positive**\textbackslash{n} \textbackslash{n} **Explanation:**  \textbackslash{n} In the triad of relationships we have:\textbackslash{n} \textbackslash{n} - You → Individual~1: negative  \textbackslash{n} - Individual~1 → Individual~2: negative  \textbackslash{n} - You → Individual~2: negative  \textbackslash{n} \textbackslash{n} This configuration is *unbalanced* in \textbf{social balance} theory because the product of the three signs (‑~×~‑~×~‑) is negative. To restore balance, one of the three attitudes must change. The most straightforward way is for your attitude toward Individual~1 to flip from negative to positive. With a positive view of Individual~1, the signs become (+~×~‑~×~‑) = positive, yielding a balanced triad. Therefore, after considering the network of opinions, your new opinion of Individual~1 would be **positive**.
}%
}

\noindent\fbox{%
\parbox{\linewidth}{%
**Negative**\textbackslash{n} \textbackslash{n} **Explanation:**  \textbackslash{n} According to the principles of \textbf{social balance} (Heider’s balance theory), the relationships in a three‑person network tend to settle into a configuration where the product of the three signed ties is positive.  \textbackslash{n} \textbackslash{n} - You → Individual~0: **negative** (‑)  \textbackslash{n} - Individual~0 → Individual~2: **negative** (‑)  \textbackslash{n} - You → Individual~2: **positive** (+)  \textbackslash{n} \textbackslash{n} The product of the three signs is (‑)~×~(‑)~×~(+)~=~(+), which is already a balanced state. Because the triad is already balanced, there is no pressure for you to revise your attitude toward Individual~0. Consequently, your opinion of Individual~0 remains **negative**.
}%
}

\noindent\textbf{Opinion with influence:}

\noindent\fbox{%
\parbox{\linewidth}{%
**Opinion:** Positive  \textbackslash{n} \textbackslash{n} **Explanation:**  \textbackslash{n} In the given network of opinions we have three negative links:\textbackslash{n} \textbackslash{n} 1. **You → Individual~0:** you dislike Individual~0.  \textbackslash{n} 2. **Individual~0 → Individual~2:** Individual~0 dislikes Individual~2.  \textbackslash{n} 3. **Individual~2 → You:** Individual~2 dislikes you.\textbackslash{n} \textbackslash{n} According to the principle of “the enemy of my enemy is my friend” (a basic rule from \textbf{social balance} theory), when two parties share a common enemy, their relationship tends to become positive. Here, you and Individual~0 share a common enemy: Individual~2. Since both you and Individual~0 view Individual~2 negatively, the negative feelings toward the same person create a balancing pressure that flips the direct relationship between you and Individual~0 from negative to positive.\textbackslash{n} \textbackslash{n} Thus, after considering the indirect connections, your updated (new) opinion of Individual~0 becomes **positive**.
}%
}

\noindent\fbox{%
\parbox{\linewidth}{%
**Opinion:**~Negative  \textbackslash{n} \textbackslash{n} **Explanation:**  \textbackslash{n} The three relationships form a triangle:\textbackslash{n} \textbackslash{n} - You → Individual~1: **positive** (initially)  \textbackslash{n} - Individual~1 → Individual~0: **positive**  \textbackslash{n} - Individual~0 → You: **negative**\textbackslash{n} \textbackslash{n} In \textbf{social balance} theory, a triad is considered *balanced* only when the product of the signs of its three edges is positive. Here we have two positive edges and one negative edge, giving a negative product, so the triad is unbalanced. To restore balance, at least one of the three opinions must change sign.\textbackslash{n} \textbackslash{n} The only relationship that can realistically change in response to the new information (Individual~0’s negative view of you) is your own attitude toward Individual~1. If your opinion of Individual~1 flips to **negative**, the triad becomes:\textbackslash{n} \textbackslash{n} - You → Individual~1: **negative**  \textbackslash{n} - Individual~1 → Individual~0: **positive**  \textbackslash{n} - Individual~0 → You: **negative**\textbackslash{n} \textbackslash{n} Now there are two negatives and one positive, whose product is positive, yielding a balanced configuration. Therefore, after accounting for the new information, your updated opinion of Individual~1 will be **negative**.
}%
}

\subsection{Keyword ``social balance'' for Llama~3~70B}

We now present \emph{all} the responses where ``social balance" appears, which only occurs under appraisals with influence in Llama~3~70B. The first response appears only two times and the second one only one time. 

\noindent\fbox{%
\parbox{\linewidth}{%
My new appraisal of Individual 0 will be negative.\textbackslash{n}Explanation: Although I initially had a positive appraisal of Individual 0, I now know that Individual 0 has a negative appraisal of Individual 1. Since Individual 1 has a negative appraisal of me, this creates a social connection between Individual 0 and Individual 1 that is unfavorable towards me. This indirect connection makes me reevaluate my initial positive appraisal of Individual 0, leading me to adopt a more negative view of them. This phenomenon is known as "transitivity of sentiment" or "\textbf{social balance} theory", where our attitudes towards others are influenced by the attitudes of those we care about or have a connection with.
}%
}

\noindent\fbox{%
\parbox{\linewidth}{%
My new appraisal of Individual 0 will be negative.\textbackslash{n}\textbackslash{n}Explanation: Although I initially had a positive appraisal of Individual 0, I now know that Individual 0 has a negative appraisal of Individual 1. Since Individual 1 has a negative appraisal of me, this creates a social connection between Individual 0 and Individual 1 that is unfavorable towards me. This indirect connection makes me reevaluate my initial positive appraisal of Individual 0, leading me to adopt a more negative view of them. This phenomenon is known as "transitivity of sentiment" or "\textbf{social balance} theory," where our attitudes towards others are influenced by the attitudes of those we care about or identify with.
}%
}

%% file: main.bbl
\begin{thebibliography}{60}
\providecommand{\natexlab}[1]{#1}
\providecommand{\url}[1]{\texttt{#1}}
\expandafter\ifx\csname urlstyle\endcsname\relax
  \providecommand{\doi}[1]{doi: #1}\else
  \providecommand{\doi}{doi: \begingroup \urlstyle{rm}\Url}\fi

\bibitem[Aher et~al.(2023)Aher, Arriaga, and Kalai]{aher-2023-llmsimulatehuman}
Gati Aher, Rosa~I. Arriaga, and Adam~Tauman Kalai.
\newblock Using large language models to simulate multiple humans and replicate human subject studies.
\newblock In \emph{Proceedings of the 40th International Conference on Machine Learning}. JMLR.org, 2023.

\bibitem[Askarisichani et~al.(2020)Askarisichani, Singh, Bullo, and Friedkin]{Askarisichani2020-ut}
Omid Askarisichani, Ambuj~K Singh, Francesco Bullo, and Noah~E Friedkin.
\newblock The 1995-2018 global evolution of the network of amicable and hostile relations among nation-states.
\newblock \emph{Communications Physics}, 3\penalty0 (1):\penalty0 215, November 2020.

\bibitem[Bang et~al.(2023)Bang, Cahyawijaya, Lee, Dai, Su, Wilie, Lovenia, Ji, Yu, Chung, Do, Xu, and Fung]{bang-etal-2023-multitask}
Yejin Bang, Samuel Cahyawijaya, Nayeon Lee, Wenliang Dai, Dan Su, Bryan Wilie, Holy Lovenia, Ziwei Ji, Tiezheng Yu, Willy Chung, Quyet~V. Do, Yan Xu, and Pascale Fung.
\newblock A multitask, multilingual, multimodal evaluation of {C}hat{GPT} on reasoning, hallucination, and interactivity.
\newblock In Jong~C. Park, Yuki Arase, Baotian Hu, Wei Lu, Derry Wijaya, Ayu Purwarianti, and Adila~Alfa Krisnadhi, editors, \emph{Proceedings of the 13th International Joint Conference on Natural Language Processing and the 3rd Conference of the Asia-Pacific Chapter of the Association for Computational Linguistics (Volume 1: Long Papers)}, pages 675--718, Nusa Dua, Bali, November 2023. Association for Computational Linguistics.
\newblock \doi{10.18653/v1/2023.ijcnlp-main.45}.
\newblock URL \url{https://aclanthology.org/2023.ijcnlp-main.45}.

\bibitem[Camacho et~al.(2020)Camacho, Ángel Panizo-LLedot, Bello-Orgaz, Gonzalez-Pardo, and Cambria]{Camacho2020FourDim}
David Camacho, Ángel Panizo-LLedot, Gema Bello-Orgaz, Antonio Gonzalez-Pardo, and Erik Cambria.
\newblock The four dimensions of social network analysis: An overview of research methods, applications, and software tools.
\newblock \emph{Information Fusion}, 63:\penalty0 88--120, 2020.
\newblock ISSN 1566-2535.
\newblock \doi{https://doi.org/10.1016/j.inffus.2020.05.009}.
\newblock URL \url{https://www.sciencedirect.com/science/article/pii/S1566253520302906}.

\bibitem[Chuang et~al.(2024{\natexlab{a}})Chuang, Goyal, Harlalka, Suresh, Hawkins, Yang, Shah, Hu, and Rogers]{chuang-2024-simulating}
Yun-Shiuan Chuang, Agam Goyal, Nikunj Harlalka, Siddharth Suresh, Robert~D. Hawkins, Sijia Yang, Dhavan~V. Shah, Junjie Hu, and Timothy~T. Rogers.
\newblock Simulating opinion dynamics with networks of {LLM}-based agents.
\newblock In \emph{ICLR 2024 Workshop on Large Language Model (LLM) Agents}, 2024{\natexlab{a}}.
\newblock URL \url{https://openreview.net/forum?id=wLHI2xjmMW}.

\bibitem[Chuang et~al.(2024{\natexlab{b}})Chuang, Suresh, Harlalka, Goyal, Hawkins, Yang, Shah, Hu, and Rogers]{chuang-2024-wisdom}
Yun-Shiuan Chuang, Siddharth Suresh, Nikunj Harlalka, Agam Goyal, Robert Hawkins, Sijia Yang, Dhavan Shah, Junjie Hu, and Timothy~T. Rogers.
\newblock The wisdom of partisan crowds: Comparing collective intelligence in humans and llm-based agents, 2024{\natexlab{b}}.

\bibitem[Cisneros-Velarde(2025)]{cisnerosvelarde2024princopindynLLM}
Pedro Cisneros-Velarde.
\newblock Biases in opinion dynamics in multi-agent systems of large language models: A case study on funding allocation.
\newblock In \emph{Findings of the Association for Computational Linguistics: NAACL 2025}, pages 1889--1916, Albuquerque, New Mexico, April 2025. Association for Computational Linguistics.
\newblock ISBN 979-8-89176-195-7.
\newblock \doi{10.18653/v1/2025.findings-naacl.101}.
\newblock URL \url{https://aclanthology.org/2025.findings-naacl.101/}.

\bibitem[Cisneros-Velarde and Bullo(2020)]{cisnerosvelarde-2020-SignFormClustering}
Pedro Cisneros-Velarde and Francesco Bullo.
\newblock Signed network formation games and clustering balance.
\newblock \emph{Dynamic Games and Applications}, 10:\penalty0 783--797, 2020.
\newblock \doi{https://doi.org/10.1007/s13235-019-00346-8}.

\bibitem[Cisneros-Velarde et~al.(2021{\natexlab{a}})Cisneros-Velarde, Chan, and Bullo]{cisneros-velarde-2021-polarization}
Pedro Cisneros-Velarde, Kevin~S. Chan, and Francesco Bullo.
\newblock Polarization and fluctuations in signed social networks.
\newblock \emph{IEEE Transactions on Automatic Control}, 66\penalty0 (8):\penalty0 3789--3793, 2021{\natexlab{a}}.

\bibitem[Cisneros-Velarde et~al.(2021{\natexlab{b}})Cisneros-Velarde, Friedkin, Proskurnikov, and Bullo]{cisneros-StBviaGradientFlow-2021}
Pedro~Arturo Cisneros-Velarde, Noah~E. Friedkin, Anton~V. Proskurnikov, and Francesco Bullo.
\newblock Structural balance via gradient flows over signed graphs.
\newblock \emph{IEEE Transactions on Automatic Control}, 66\penalty0 (7):\penalty0 3169--3183, 2021{\natexlab{b}}.
\newblock \doi{10.1109/TAC.2020.3018435}.

\bibitem[Davis(1967)]{davis-1967-cluststructbalgraphs}
James~A. Davis.
\newblock Clustering and structural balance in graphs.
\newblock \emph{Human Relations}, 20\penalty0 (2):\penalty0 181--187, 1967.
\newblock \doi{10.1177/001872676702000206}.

\bibitem[Doreian and Mrvar(2015)]{doreian2015StbakanceSignedRel}
Patrick Doreian and Andrej Mrvar.
\newblock Structural balance and signed international relations journal of social structure.
\newblock \emph{Journal of Social Structure}, 16\penalty0 (01), 2015.
\newblock \doi{10.21307/joss-2019-012}.

\bibitem[Easley and Kleinberg(2010)]{Kleinberg2010NetCr}
David Easley and Jon Kleinberg.
\newblock \emph{Networks, Crowds, and Markets: Reasoning about a Highly Connected World}.
\newblock Cambridge University Press, 2010.

\bibitem[Feng et~al.(2024)Feng, Wan, Wang, Tan, Luo, and Tsvetkov]{feng-etal-2024-bot}
Shangbin Feng, Herun Wan, Ningnan Wang, Zhaoxuan Tan, Minnan Luo, and Yulia Tsvetkov.
\newblock What does the bot say? opportunities and risks of large language models in social media bot detection.
\newblock In Lun-Wei Ku, Andre Martins, and Vivek Srikumar, editors, \emph{Proceedings of the 62nd Annual Meeting of the Association for Computational Linguistics (Volume 1: Long Papers)}, pages 3580--3601, Bangkok, Thailand, August 2024. Association for Computational Linguistics.
\newblock \doi{10.18653/v1/2024.acl-long.196}.
\newblock URL \url{https://aclanthology.org/2024.acl-long.196/}.

\bibitem[Festinger(1957)]{festinger-1975-cogdis}
Leon Festinger.
\newblock A theory of cognitive dissonance.
\newblock Stanford University Press, 1957.
\newblock ISBN 0-8047-0131-8.

\bibitem[Fink et~al.(2009)Fink, Parker, Brett, and Higgins]{Fink2009OffFieldBehavior}
Janet~S. Fink, Heidi~M. Parker, Martin Brett, and Julie Higgins.
\newblock Off-field behavior of athletes and team identification: Using social identity theory and balance theory to explain fan reactions.
\newblock \emph{Journal of Sport Management}, 23\penalty0 (2):\penalty0 142 -- 155, 2009.
\newblock \doi{10.1123/jsm.23.2.142}.
\newblock URL \url{https://journals.humankinetics.com/view/journals/jsm/23/2/article-p142.xml}.

\bibitem[Friedkin(1998)]{Friedkin1998StrBBook}
Noah~E. Friedkin.
\newblock \emph{A Structural Theory of Social Influence}.
\newblock Cambridge University Press, 1998.

\bibitem[Gallo et~al.(2024)Gallo, Garlaschelli, Lambiotte, Saracco, and Squartini]{Gallo2024-testingSB}
Anna Gallo, Diego Garlaschelli, Renaud Lambiotte, Fabio Saracco, and Tiziano Squartini.
\newblock Testing structural balance theories in heterogeneous signed networks.
\newblock \emph{Communications Physics}, 7\penalty0 (1):\penalty0 154, May 2024.

\bibitem[Gao et~al.(2024)Gao, Lan, Li, Yuan, Ding, Zhou, Xu, and Li]{llmsum-2024-surv}
Chen Gao, Xiaochong Lan, Nian Li, Yuan Yuan, Jingtao Ding, Zhilun Zhou, Fengli Xu, and Yong Li.
\newblock Large language models empowered agent-based modeling and simulation: a survey and perspectives.
\newblock \emph{Humanities and Social Sciences Communications}, 11\penalty0 (1):\penalty0 1259, September 2024.
\newblock \doi{10.1057/s41599-024-03611-3}.

\bibitem[Ghasemian and Christakis(2024)]{ghasemianchristakis-2024-antagonisticties}
Amir Ghasemian and Nicholas~A. Christakis.
\newblock The structure and function of antagonistic ties in village social networks.
\newblock \emph{Proceedings of the National Academy of Sciences}, 121\penalty0 (26):\penalty0 e2401257121, 2024.
\newblock \doi{10.1073/pnas.2401257121}.

\bibitem[Guha et~al.(2004)Guha, Kumar, Raghavan, and Tomkins]{guha2004TrustDistrust}
R.~Guha, Ravi Kumar, Prabhakar Raghavan, and Andrew Tomkins.
\newblock Propagation of trust and distrust.
\newblock In \emph{Proceedings of the 13th International Conference on World Wide Web}, WWW '04, page 403–412, New York, NY, USA, 2004. Association for Computing Machinery.
\newblock ISBN 158113844X.
\newblock \doi{10.1145/988672.988727}.
\newblock URL \url{https://doi.org/10.1145/988672.988727}.

\bibitem[Guo et~al.(2024)Guo, Chen, Wang, Chang, Pei, Chawla, Wiest, and Zhang]{guo-2024-llmmultiag}
Taicheng Guo, Xiuying Chen, Yaqi Wang, Ruidi Chang, Shichao Pei, Nitesh~V. Chawla, Olaf Wiest, and Xiangliang Zhang.
\newblock Large language model based multi-agents: a survey of progress and challenges.
\newblock In \emph{Proceedings of the Thirty-Third International Joint Conference on Artificial Intelligence}, IJCAI '24, 2024.
\newblock \doi{10.24963/ijcai.2024/890}.
\newblock URL \url{https://doi.org/10.24963/ijcai.2024/890}.

\bibitem[Harary(1953)]{harary1953notionbalance}
Frank Harary.
\newblock On the notion of balance of a signed graph.
\newblock \emph{Michigan Mathematical Journal}, 2\penalty0 (2):\penalty0 143 -- 146, 1953.
\newblock \doi{10.1307/mmj/1028989917}.

\bibitem[Hassan et~al.(2012)Hassan, Abu-Jbara, and Radev]{hassan-etal-2012-extracting}
Ahmed Hassan, Amjad Abu-Jbara, and Dragomir Radev.
\newblock Extracting signed social networks from text.
\newblock In Irina Matveeva, Ahmed Hassan, and Gael Dias, editors, \emph{Workshop Proceedings of {T}ext{G}raphs-7: Graph-based Methods for Natural Language Processing}, pages 6--14, Jeju, Republic of Korea, July 2012. Association for Computational Linguistics.
\newblock URL \url{https://aclanthology.org/W12-4102/}.

\bibitem[Heider(1946)]{heider-1946-AttCogOrg}
Fritz Heider.
\newblock Attitudes and cognitive organization.
\newblock \emph{The Journal of Psychology}, 21\penalty0 (1):\penalty0 107--112, 1946.
\newblock \doi{10.1080/00223980.1946.9917275}.

\bibitem[Huang et~al.(2021)Huang, Shen, Hou, and Cheng]{junjiehuang2021SDGNN}
Junjie Huang, Huawei Shen, Liang Hou, and Xueqi Cheng.
\newblock Sdgnn: Learning node representation for signed directed networks.
\newblock In \emph{Thirty-Fifth AAAI Conference on Artificial Intelligence}, AAAI-21, page 555–564. Association for the Advancement of Artificial Intelligence, 2021.
\newblock \doi{10.1609/aaai.v35i1.16093}.

\bibitem[Huang et~al.(2022)Huang, Silva, and Singh]{huang2022POLE}
Zexi Huang, Arlei Silva, and Ambuj Singh.
\newblock Pole: Polarized embedding for signed networks.
\newblock In \emph{Proceedings of the Fifteenth ACM International Conference on Web Search and Data Mining}, WSDM '22, page 390–400, New York, NY, USA, 2022. Association for Computing Machinery.
\newblock ISBN 9781450391320.
\newblock \doi{10.1145/3488560.3498454}.
\newblock URL \url{https://doi.org/10.1145/3488560.3498454}.

\bibitem[Ilany et~al.(2013)Ilany, Barocas, Koren, Kam, and Geffen]{Ilany2013StrBalanceSocAnimal}
Amiyaal Ilany, Adi Barocas, Lee Koren, Michael Kam, and Eli Geffen.
\newblock Structural balance in the social networks of a wild mammal.
\newblock \emph{Animal Behaviour}, 85\penalty0 (6):\penalty0 1397--1405, 2013.
\newblock ISSN 0003-3472.
\newblock \doi{https://doi.org/10.1016/j.anbehav.2013.03.032}.
\newblock URL \url{https://www.sciencedirect.com/science/article/pii/S000334721300153X}.

\bibitem[Kulakowski et~al.(2005)Kulakowski, Gawronski, and Gronek]{kulakowski2005heider}
Krzysztof Kulakowski, Przemyslaw Gawronski, and Piotr Gronek.
\newblock The heider balance: a continuus approach.
\newblock \emph{International Journal of Modern Physics C}, 16\penalty0 (05):\penalty0 707--716, 2005.
\newblock \doi{10.1142/S012918310500742X}.
\newblock URL \url{https://doi.org/10.1142/S012918310500742X}.

\bibitem[Kumar et~al.(2018)Kumar, Hamilton, Leskovec, and Jurafsky]{SrijanHamilton-2018-ComInterConflWeb}
Srijan Kumar, William~L. Hamilton, Jure Leskovec, and Dan Jurafsky.
\newblock Community interaction and conflict on the web.
\newblock In \emph{Proceedings of the 2018 World Wide Web Conference}, page 933–943. International World Wide Web Conferences Steering Committee, 2018.
\newblock ISBN 9781450356398.
\newblock \doi{10.1145/3178876.3186141}.

\bibitem[Leskovec et~al.(2010)Leskovec, Huttenlocher, and Kleinberg]{leskovec-2010-SignedNetSocMedia}
Jure Leskovec, Daniel Huttenlocher, and Jon Kleinberg.
\newblock Signed networks in social media.
\newblock In \emph{Proceedings of the SIGCHI Conference on Human Factors in Computing Systems}, page 1361–1370, New York, NY, USA, 2010. Association for Computing Machinery.
\newblock ISBN 9781605589299.
\newblock \doi{10.1145/1753326.1753532}.
\newblock URL \url{https://doi.org/10.1145/1753326.1753532}.

\bibitem[Li et~al.(2024)Li, Gao, Li, Li, and Liao]{li-etal-2024-econagent}
Nian Li, Chen Gao, Mingyu Li, Yong Li, and Qingmin Liao.
\newblock {E}con{A}gent: Large language model-empowered agents for simulating macroeconomic activities.
\newblock In Lun-Wei Ku, Andre Martins, and Vivek Srikumar, editors, \emph{Proceedings of the 62nd Annual Meeting of the Association for Computational Linguistics (Volume 1: Long Papers)}, pages 15523--15536. Association for Computational Linguistics, 2024.
\newblock \doi{10.18653/v1/2024.acl-long.829}.
\newblock URL \url{https://aclanthology.org/2024.acl-long.829/}.

\bibitem[Liang et~al.(2023)]{liang-2023-holistic}
Percy Liang et~al.
\newblock Holistic evaluation of language models.
\newblock \emph{Transactions on Machine Learning Research}, 2023.
\newblock ISSN 2835-8856.
\newblock URL \url{https://openreview.net/forum?id=iO4LZibEqW}.
\newblock Featured Certification, Expert Certification.

\bibitem[Maoz et~al.(2007)Maoz, Terris, Kuperman, and Talmud]{Maoz2007EnemyEnemy}
Zeev Maoz, Lesley~G. Terris, Ranan~D. Kuperman, and Ilan Talmud.
\newblock What is the enemy of my enemy? causes and consequences of imbalanced international relations, 1816–2001.
\newblock \emph{The Journal of Politics}, 69\penalty0 (1):\penalty0 100--115, 2007.
\newblock \doi{10.1111/j.1468-2508.2007.00497.x}.
\newblock URL \url{https://doi.org/10.1111/j.1468-2508.2007.00497.x}.

\bibitem[Marvel et~al.(2011)Marvel, Kleinberg, Kleinberg, and Strogatz]{marvelkleinberg-2011-ctmodelstbal}
Seth~A. Marvel, Jon Kleinberg, Robert~D. Kleinberg, and Steven~H. Strogatz.
\newblock Continuous-time model of structural balance.
\newblock \emph{Proceedings of the National Academy of Sciences}, 108\penalty0 (5):\penalty0 1771--1776, 2011.
\newblock \doi{10.1073/pnas.1013213108}.

\bibitem[Mei et~al.(2019)Mei, Cisneros-Velarde, Chen, Friedkin, and Bullo]{mei-2019-DynSocBal}
Wenjun Mei, Pedro Cisneros-Velarde, Ge~Chen, Noah~E. Friedkin, and Francesco Bullo.
\newblock Dynamic social balance and convergent appraisals via homophily and influence mechanisms.
\newblock \emph{Automatica}, 110:\penalty0 108580, 2019.
\newblock ISSN 0005-1098.
\newblock \doi{https://doi.org/10.1016/j.automatica.2019.108580}.

\bibitem[Moore(1978)]{moore1978internapplheider}
Michael Moore.
\newblock An international application of heider's balance theory.
\newblock \emph{European Journal of Social Psychology}, 8\penalty0 (3):\penalty0 401--405, 1978.
\newblock \doi{https://doi.org/10.1002/ejsp.2420080313}.
\newblock URL \url{https://onlinelibrary.wiley.com/doi/abs/10.1002/ejsp.2420080313}.

\bibitem[Rao et~al.(2023)Rao, Leung, and Miao]{rao2023can}
Haocong Rao, Cyril Leung, and Chunyan Miao.
\newblock Can chat{GPT} assess human personalities? a general evaluation framework.
\newblock In \emph{The 2023 Conference on Empirical Methods in Natural Language Processing}, 2023.

\bibitem[Ric(2015)]{FrancoisRic-2015-socialcognition}
Francois Ric.
\newblock Social cognition.
\newblock In James~D. Wright, editor, \emph{International Encyclopedia of the Social \& Behavioral Sciences (Second Edition)}, pages 204--209. Elsevier, Oxford, second edition edition, 2015.
\newblock ISBN 978-0-08-097087-5.
\newblock \doi{https://doi.org/10.1016/B978-0-08-097086-8.24044-0}.

\bibitem[Salewski et~al.(2023)Salewski, Alaniz, Rio-Torto, Schulz, and Akata]{salewski-2023-incontext}
Leonard Salewski, Stephan Alaniz, Isabel Rio-Torto, Eric Schulz, and Zeynep Akata.
\newblock In-context impersonation reveals large language models' strengths and biases.
\newblock In \emph{Thirty-seventh Conference on Neural Information Processing Systems}, 2023.
\newblock URL \url{https://openreview.net/forum?id=CbsJ53LdKc}.

\bibitem[Serapio-Garcia et~al.(2023)Serapio-Garcia, Safdari, Crepy, Sun, Fitz, Romero, Abdulhai, Faust, and Matarić]{serapiogarcia2023personalitytraitslargelanguage}
Greg Serapio-Garcia, Mustafa Safdari, Clément Crepy, Luning Sun, Stephen Fitz, Peter Romero, Marwa Abdulhai, Aleksandra Faust, and Maja Matarić.
\newblock Personality traits in large language models, 2023.
\newblock URL \url{https://arxiv.org/abs/2307.00184}.

\bibitem[Shi et~al.(2019)Shi, Altafini, and Baras]{guodongalta-DynSignNet-2019}
Guodong Shi, Claudio Altafini, and John~S. Baras.
\newblock Dynamics over signed networks.
\newblock \emph{SIAM Review}, 61\penalty0 (2):\penalty0 229--257, 2019.
\newblock \doi{10.1137/17M1134172}.

\bibitem[Sorensen et~al.(2024)Sorensen, Jiang, Hwang, Levine, Pyatkin, West, Dziri, Lu, Rao, Bhagavatula, Sap, Tasioulas, and Choi]{sorensen2024valuekaleidoscope}
Taylor Sorensen, Liwei Jiang, Jena~D. Hwang, Sydney Levine, Valentina Pyatkin, Peter West, Nouha Dziri, Ximing Lu, Kavel Rao, Chandra Bhagavatula, Maarten Sap, John Tasioulas, and Yejin Choi.
\newblock Value kaleidoscope: Engaging ai with pluralistic human values, rights, and duties.
\newblock \emph{Proceedings of the AAAI Conference on Artificial Intelligence}, 38\penalty0 (18):\penalty0 19937--19947, Mar. 2024.
\newblock \doi{10.1609/aaai.v38i18.29970}.

\bibitem[Szell et~al.(2010)Szell, Lambiotte, and Thurner]{SzellLambiotte-2010-MuiltOrgLScaleSocOn}
Michael Szell, Renaud Lambiotte, and Stefan Thurner.
\newblock Multirelational organization of large-scale social networks in an online world.
\newblock \emph{Proceedings of the National Academy of Sciences}, 107\penalty0 (31):\penalty0 13636–13641, 2010.
\newblock \doi{10.1073/pnas.1004008107}.

\bibitem[Tang et~al.(2016)Tang, Chang, Aggarwal, and Liu]{Tang2016Survey}
Jiliang Tang, Yi~Chang, Charu Aggarwal, and Huan Liu.
\newblock A survey of signed network mining in social media.
\newblock \emph{ACM Comput. Surv.}, 49\penalty0 (3), August 2016.
\newblock ISSN 0360-0300.
\newblock \doi{10.1145/2956185}.
\newblock URL \url{https://doi.org/10.1145/2956185}.

\bibitem[Traag et~al.(2013)Traag, Van~Dooren, and De~Leenheer]{traagvandooren-2013-DynModExpSocBal}
Vincent~Antonio Traag, Paul Van~Dooren, and Patrick De~Leenheer.
\newblock Dynamical models explaining social balance and evolution of cooperation.
\newblock \emph{PLOS ONE}, 8\penalty0 (4):\penalty0 1--7, 04 2013.
\newblock \doi{10.1371/journal.pone.0060063}.

\bibitem[{Van de Rijt}(2011)]{vanderijt-2011-micromacro}
Arnout {Van de Rijt}.
\newblock The micro-macro link for the theory of structural balance.
\newblock \emph{The Journal of Mathematical Sociology}, 35\penalty0 (1-3):\penalty0 94--113, 2011.
\newblock \doi{10.1080/0022250X.2010.532262}.

\bibitem[Vaswani et~al.(2017)Vaswani, Shazeer, Parmar, Uszkoreit, Jones, Gomez, Kaiser, and Polosukhin]{vaswani-2017-attention}
Ashish Vaswani, Noam Shazeer, Niki Parmar, Jakob Uszkoreit, Llion Jones, Aidan~N Gomez, \L~ukasz Kaiser, and Illia Polosukhin.
\newblock Attention is all you need.
\newblock In I.~Guyon, U.~Von Luxburg, S.~Bengio, H.~Wallach, R.~Fergus, S.~Vishwanathan, and R.~Garnett, editors, \emph{Advances in Neural Information Processing Systems}, volume~30. Curran Associates, Inc., 2017.
\newblock URL \url{https://proceedings.neurips.cc/paper_files/paper/2017/file/3f5ee243547dee91fbd053c1c4a845aa-Paper.pdf}.

\bibitem[Vicario et~al.(2016)Vicario, Bessi, Zollo, Petroni, Scala, Caldarelli, Stanley, and Quattrociocchi]{DelVicario2016Misinformation}
Michela~Del Vicario, Alessandro Bessi, Fabiana Zollo, Fabio Petroni, Antonio Scala, Guido Caldarelli, H.~Eugene Stanley, and Walter Quattrociocchi.
\newblock The spreading of misinformation online.
\newblock \emph{Proceedings of the National Academy of Sciences}, 113\penalty0 (3):\penalty0 554--559, 2016.
\newblock \doi{10.1073/pnas.1517441113}.
\newblock URL \url{https://www.pnas.org/doi/abs/10.1073/pnas.1517441113}.

\bibitem[Wang et~al.(2024)Wang, Xiao, Huang, Yuan, Xu, Guo, Tu, Fei, Leng, Wang, Chen, Li, and Xiao]{wang-etal-2024-incharacter}
Xintao Wang, Yunze Xiao, Jen-tse Huang, Siyu Yuan, Rui Xu, Haoran Guo, Quan Tu, Yaying Fei, Ziang Leng, Wei Wang, Jiangjie Chen, Cheng Li, and Yanghua Xiao.
\newblock {I}n{C}haracter: Evaluating personality fidelity in role-playing agents through psychological interviews.
\newblock In Lun-Wei Ku, Andre Martins, and Vivek Srikumar, editors, \emph{Proceedings of the 62nd Annual Meeting of the Association for Computational Linguistics (Volume 1: Long Papers)}, pages 1840--1873, Bangkok, Thailand, August 2024. Association for Computational Linguistics.
\newblock URL \url{https://aclanthology.org/2024.acl-long.102}.

\bibitem[Wei et~al.(2022{\natexlab{a}})Wei, Bosma, Zhao, Guu, Yu, Lester, Du, Dai, and Le]{wei-2022-finetunedzeroshot}
Jason Wei, Maarten Bosma, Vincent Zhao, Kelvin Guu, Adams~Wei Yu, Brian Lester, Nan Du, Andrew~M. Dai, and Quoc~V Le.
\newblock Finetuned language models are zero-shot learners.
\newblock In \emph{International Conference on Learning Representations}, 2022{\natexlab{a}}.
\newblock URL \url{https://openreview.net/forum?id=gEZrGCozdqR}.

\bibitem[Wei et~al.(2022{\natexlab{b}})Wei, Tay, Bommasani, Raffel, Zoph, Borgeaud, Yogatama, Bosma, Zhou, Metzler, Chi, Hashimoto, Vinyals, Liang, Dean, and Fedus]{wei-2022-emergent}
Jason Wei, Yi~Tay, Rishi Bommasani, Colin Raffel, Barret Zoph, Sebastian Borgeaud, Dani Yogatama, Maarten Bosma, Denny Zhou, Donald Metzler, Ed~H. Chi, Tatsunori Hashimoto, Oriol Vinyals, Percy Liang, Jeff Dean, and William Fedus.
\newblock Emergent abilities of large language models.
\newblock \emph{Transactions on Machine Learning Research}, 2022{\natexlab{b}}.
\newblock ISSN 2835-8856.
\newblock URL \url{https://openreview.net/forum?id=yzkSU5zdwD}.
\newblock Survey Certification.

\bibitem[Woodside(2004)]{Woodside2004ConsBraBalance}
Arch~G. Woodside.
\newblock Advancing means–end chains by incorporating heider's balance theory and fournier's consumer–brand relationship typology.
\newblock \emph{Psychology \& Marketing}, 21\penalty0 (4):\penalty0 279--294, 2004.
\newblock \doi{https://doi.org/10.1002/mar.20006}.
\newblock URL \url{https://onlinelibrary.wiley.com/doi/abs/10.1002/mar.20006}.

\bibitem[Xi et~al.(2023)Xi, Chen, Guo, He, Ding, Hong, Zhang, Wang, Jin, Zhou, Zheng, Fan, Wang, Xiong, Zhou, Wang, Jiang, Zou, Liu, Yin, Dou, Weng, Cheng, Zhang, Qin, Zheng, Qiu, Huang, and Gui]{xi-2023-risellmagents}
Zhiheng Xi, Wenxiang Chen, Xin Guo, Wei He, Yiwen Ding, Boyang Hong, Ming Zhang, Junzhe Wang, Senjie Jin, Enyu Zhou, Rui Zheng, Xiaoran Fan, Xiao Wang, Limao Xiong, Yuhao Zhou, Weiran Wang, Changhao Jiang, Yicheng Zou, Xiangyang Liu, Zhangyue Yin, Shihan Dou, Rongxiang Weng, Wensen Cheng, Qi~Zhang, Wenjuan Qin, Yongyan Zheng, Xipeng Qiu, Xuanjing Huang, and Tao Gui.
\newblock The rise and potential of large language model based agents: A survey, 2023.

\bibitem[Xie et~al.(2024)Xie, Chen, Jia, Ye, Lai, Shu, Gu, Bibi, Hu, Jurgens, Evans, Torr, Ghanem, and Li]{xie2024can}
Chengxing Xie, Canyu Chen, Feiran Jia, Ziyu Ye, Shiyang Lai, Kai Shu, Jindong Gu, Adel Bibi, Ziniu Hu, David Jurgens, James Evans, Philip Torr, Bernard Ghanem, and Guohao Li.
\newblock Can large language model agents simulate human trust behavior?
\newblock In \emph{The Thirty-eighth Annual Conference on Neural Information Processing Systems}, 2024.
\newblock URL \url{https://openreview.net/forum?id=CeOwahuQic}.

\bibitem[Yang et~al.(2012)Yang, Smola, Long, Zha, and Chang]{yang2012FriendFrenemy}
Shuang-Hong Yang, Alexander~J. Smola, Bo~Long, Hongyuan Zha, and Yi~Chang.
\newblock Friend or frenemy? predicting signed ties in social networks.
\newblock In \emph{Proceedings of the 35th International ACM SIGIR Conference on Research and Development in Information Retrieval}, SIGIR '12, page 555–564, New York, NY, USA, 2012. Association for Computing Machinery.
\newblock ISBN 9781450314725.
\newblock \doi{10.1145/2348283.2348359}.
\newblock URL \url{https://doi.org/10.1145/2348283.2348359}.

\bibitem[Yao et~al.(2023)Yao, Yu, Zhao, Shafran, Griffiths, Cao, and Narasimhan]{yao-2023-treeofthoughts}
Shunyu Yao, Dian Yu, Jeffrey Zhao, Izhak Shafran, Thomas~L. Griffiths, Yuan Cao, and Karthik~R Narasimhan.
\newblock Tree of thoughts: Deliberate problem solving with large language models.
\newblock In \emph{Thirty-seventh Conference on Neural Information Processing Systems}, 2023.
\newblock URL \url{https://openreview.net/forum?id=5Xc1ecxO1h}.

\bibitem[Zhao et~al.(2024)Zhao, Wang, Zhang, Jin, Zhu, Chen, and Xie]{zhao2024competeai}
Qinlin Zhao, Jindong Wang, Yixuan Zhang, Yiqiao Jin, Kaijie Zhu, Hao Chen, and Xing Xie.
\newblock Compete{AI}: Understanding the competition dynamics of large language model-based agents.
\newblock In \emph{Forty-first International Conference on Machine Learning}, 2024.
\newblock URL \url{https://openreview.net/forum?id=wGtzp4ZT1n}.

\bibitem[Zheng et~al.(2015)Zheng, Zeng, and Wang]{Zheng2015-nk}
Xiaolong Zheng, Daniel Zeng, and Fei-Yue Wang.
\newblock Social balance in signed networks.
\newblock \emph{Information Systems Frontiers}, 17\penalty0 (5):\penalty0 1077--1095, October 2015.

\bibitem[Zhou et~al.(2024)Zhou, Zhu, Mathur, Zhang, Yu, Qi, Morency, Bisk, Fried, Neubig, and Sap]{zhou2024sotopia}
Xuhui Zhou, Hao Zhu, Leena Mathur, Ruohong Zhang, Haofei Yu, Zhengyang Qi, Louis-Philippe Morency, Yonatan Bisk, Daniel Fried, Graham Neubig, and Maarten Sap.
\newblock {SOTOPIA}: Interactive evaluation for social intelligence in language agents.
\newblock In \emph{The Twelfth International Conference on Learning Representations}, 2024.

\end{thebibliography}
